\documentclass[lettersize,journal]{IEEEtran}
\usepackage{amsmath,amsfonts}
\usepackage{algorithmic}
\usepackage{algorithm}
\usepackage{array}
\usepackage[caption=false,font=normalsize,labelfont=sf,textfont=sf]{subfig}
\usepackage{textcomp}
\usepackage{stfloats}
\usepackage{url}
\usepackage{verbatim}
\usepackage{graphicx}
\usepackage{cite}
\usepackage{hyperref}       
\usepackage{multirow}
\usepackage{multicol} 
\usepackage{amsmath, amssymb, amsfonts}
\usepackage{booktabs}  
\usepackage{array}
\usepackage{stfloats}
\usepackage{xcolor}         
\begin{document}

\title{MIDAS: Mutual Information Disentanglement with Uncertainty-Aware Fusion for Incomplete Multimodal Sentiment Analysis}

\author{Yuhua Wen, Yingying Zhou, Qifei Li, Yingming Gao,~\IEEEmembership{Member,~IEEE,} Zhengqi Wen,~\IEEEmembership{Member,~IEEE,} \\ Jianhua Tao,~\IEEEmembership{Senior Member,~IEEE,} and Ya Li,~\IEEEmembership{Member,~IEEE}

\thanks{This work is supported by the National Key R\&D Program of China under Grant No.2024YFB2808802. (Corresponding author: Ya Li.)}
\thanks{Yuhua Wen is with the School of Artificial Intelligence, Beijing University of Posts and Telecommunications, Beijing 100876, China, and also with the Zhongguancun Academy, Beijing 100094, China (e-mail: yuhuawen@bupt.edu.cn).}
\thanks{Yingying Zhou, Qifei Li, Yingming Gao, and Ya Li are with the School of Artificial Intelligence, Beijing University of Posts and Telecommunications, Beijing 100876, China (e-mail: yingyingzhou@bupt.edu.cn; liqifei@bupt.edu.cn; yingming.gao@bupt.edu.cn; yli01@bupt.edu.cn).}
\thanks{Zhengqi Wen is with the Beijing National Research Center for Information Science and Technology, Tsinghua University, Beijing 100084, China (e-mail: zqwen@tsinghua.edu.cn).}
\thanks{Jianhua Tao is with the Department of Automation, Tsinghua University, Beijing 100084, China, and also with the Beijing National Research Center for Information Science and Technology, Tsinghua University, Beijing 100084, China (e-mail: jhtao@tsinghua.edu.cn).}
\thanks{The codes are available at https://github.com/ultramarineX/MIDAS.}
}

\markboth{Journal of \LaTeX\ Class Files,~Vol.~14, No.~8, August~2021}%
{Shell \MakeLowercase{\textit{et al.}}: A Sample Article Using IEEEtran.cls for IEEE Journals}


\maketitle

\begin{abstract}
Most existing multimodal sentiment analysis approaches assume access to complete multimodal inputs. However, real-world applications frequently encounter incomplete or corrupted modalities, posing a critical challenge. Although several methods have been proposed to tackle this issue, they mainly rely on data imputation and heuristic coordination constraints, which fail to effectively extract and leverage task-relevant information from the incomplete multimodal data.
To address this challenge, we propose a unified framework termed Mutual Information Disentanglement with uncertainty-Aware fuSion (MIDAS), which effectively restructures multimodal representations under incomplete conditions.
MIDAS adopts a variational modeling strategy to represent each modality with multivariate Gaussian latent variables and further decomposes them into shared and exclusive factors. To obtain reliable representations, we design a minimax objective that minimizes the mutual information between shared and exclusive spaces for stable disentanglement, while maximizing the mutual information among shared spaces across modalities to enhance semantic alignment. In addition, an uncertainty-aware fusion mechanism is introduced, where posterior variance is leveraged as a reliability indicator to adaptively weight latent features during fusion, ensuring robust integration even when modalities are incomplete.
Extensive experiments on three widely used datasets show that MIDAS achieves strong and consistent performance gains over competitive baselines across a wide range of incomplete settings, demonstrating its effectiveness and robustness for incomplete data scenarios.
\end{abstract}

\begin{IEEEkeywords}
Multimodal Sentiment analysis, incomplete multimodal learning, mutual informantion, uncertainty-aware fusion.
\end{IEEEkeywords}

\section{Introduction}

\begin{figure}[ht]
  \centering
  \includegraphics[width=0.475\textwidth]{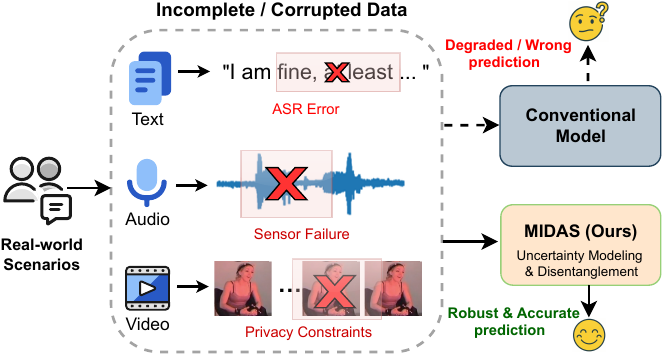}
  \caption{Illustration of incomplete multimodal sentiment analysis in real-world scenarios. 
Conventional models suffer from unstable representations under incomplete inputs, whereas MIDAS learns robust multimodal representations.}
  \label{fig:first}
\end{figure}

\IEEEPARstart{M}{ultimodal} Sentiment Analysis (MSA) aims to infer human affective states by leveraging information from multiple modalities, such as language, audio, and visual signals. By integrating complementary cues across modalities, MSA provides a more comprehensive understanding of sentiment than unimodal approaches \cite{das2023MSA, GANDHI2023MSA}. However, as illustrated in Fig.~\ref{fig:first}, real-world multimodal inputs are frequently incomplete or corrupted due to sensor failure, asynchrony, bandwidth limitations, or privacy constraints \cite{wu2024deep}. Such incomplete conditions often lead to unstable or unreliable representations in conventional models. As MSA applications continue to expand to mobile platforms \cite{wang2024review}, human-computer interaction \cite{moin2023emotion}, and edge devices \cite{zhang2023ecmer}, ensuring robustness under incomplete multimodal inputs has become imperative.

Leveraging the inherent redundancy of multimodal data, existing studies have proposed various strategies to handle incomplete multimodal inputs. These methods can be broadly categorized into two paradigms: data imputation-based methods and coordinated representation-based methods. The former aims to first reconstruct missing modalities and then proceed with traditional multimodal methods \cite{luo2023multimodal,yuan2021tfr-net,sun2023efficient,liu2024contrastive,yuan2023NIAT,wang2023distribution}. While intuitive, they incur substantial computational overhead and often suffer from error accumulation due to imperfect reconstruction. In contrast, the latter focuses on learning representations that remain semantically consistent across all available modality combinations \cite{zhao2021missing,liu2024modality,li2024UMDF,li2024correlation,zhang2024lnln}. 
However, due to intrinsic heterogeneity and discontinuities across modality feature spaces, these methods often rely on complex coordination mechanisms and still struggle to learn consistent features across modalities. More importantly, these methods fail to effectively extract and exploit the task-relevant information from the corrupted multimodal data, leading easily degrade under severely incomplete conditions.

To address these limitations, it is essential to better characterize the unique properties of incomplete MSA, which differ fundamentally from those of complete MSA.
First, incomplete multimodal inputs are often noisy, entangled, and semantically mixed, making it difficult to distinguish sentiment-relevant cues from modality-specific artifacts. Second, different missing patterns effectively induce distinct input distributions, leading to significant distribution shifts between training and inference. From this perspective, incomplete MSA closely resembles a domain generalization problem, where models must generalize across heterogeneous and partially observed domains. These challenges motivate the adoption of disentangled representation learning \cite{wang2024disentangled}, which aims to reorganize corrupted multimodal information by suppressing redundancy and noise while explicitly separating shared semantic factors from modality-specific details. Such structured representations provide a principled mechanism to enhance robustness across missing settings and alleviate distributional shifts inherent in incomplete MSA.

Although disentangled representation learning has demonstrated effectiveness in complete MSA scenarios \cite{hazarika2020MISA,yang2022FEDER}, existing approaches predominantly rely on orthogonality constraints or adversarial objectives to enforce disentanglement. These techniques impose only indirect or geometric separation and are prone to unstable optimization or degenerate solutions, particularly under incomplete inputs \cite{zhang2024lnln}. More fundamentally, disentanglement is an information separation problem, whose goal is to decompose representations into components with minimal semantic overlap. Mutual information (MI), as a foundational measure of statistical dependence, offers a natural and principled criterion for this purpose. By directly quantifying shared information between latent factors, MI-based disentanglement provides a more fundamental alternative to heuristic constraints. Moreover, MI-based objectives are less sensitive to data distributions, making them especially suitable for the diverse missing patterns encountered in incomplete MSA. However, existing MI-based methods are developed and evaluated under complete settings  \cite{sun2025multimodal}, leaving their effectiveness in incomplete scenarios insufficiently explored.

In addition to distribution shifts, incomplete multimodal inputs introduce another critical but often overlooked factor: uncertainty. When a modality is incomplete or partially observed, the reliability of its corresponding representation inevitably decreases. However, most existing methods implicitly assume uniform confidence across modalities and features, treating all representations equally during fusion \cite{zhang2024lnln,zhu2025proxy}. This mismatch leads to suboptimal or even unstable predictions, as unreliable modalities may exert disproportionate influence on the final decision. Therefore, effectively handling incomplete MSA requires not only disentangled representations but also an explicit mechanism to model and utilize the uncertainty associated with each modality.

Motivated by these considerations, we propose a unified and effective framework for incomplete MSA, termed Mutual Information Disentanglement with uncertainty-Aware fuSion (MIDAS). MIDAS consists of three core components: variational modeling, MI minimax, and uncertainty-aware fusion, which are seamlessly integrated.
Specifically, MIDAS adopts a variational modeling strategy to represent all latent variables as multivariate Gaussian distributions, enabling explicit modeling of both semantic content and modality uncertainty. Unlike prior works \cite{gao2024embracing,wei2024DMRNet}, we factorize them into shared and exclusive latent factors, establishing a principled foundation for disentangled representation learning.
Building upon this formulation, we design an information-theoretic objective for robust disentanglement. It minimizes the MI between shared and exclusive latent spaces to enforce disentanglement and filter out irrelevant noise, while maximizing the MI among shared latent spaces across modalities to enhance cross modal semantic alignment. This yields a principled and theoretically grounded mechanism for reliable representation factorization under incomplete multimodal conditions.
Furthermore, MIDAS incorporates an uncertainty-aware fusion mechanism that explicitly accounts for modality reliability without requiring an auxiliary uncertainty estimator. Specifically, it leverages the uncertainty estimates derived from variational posteriors. By using posterior variance as as a measure of confidence, the fusion module adaptively downweights unreliable modalities while emphasizing reliable ones, resulting in a more robust multimodal representation.

To verify the effectiveness of our method, extensive experiments are conducted on three benchmark MSA datasets. Quantitative results and qualitative analysis demonstrate that MIDAS consistently outperforms competitive methods under diverse incomplete settings. The main contributions are summarized as follows:

\begin{itemize}
    \item We model each modality using two multivariate normal latent distributions, enabling an explicit uncertainty-aware representation and laying the foundation for reliable mutual information optimization.
    
    \item We propose a principled mutual information disentanglement framework that jointly enforces disentanglement between shared and exclusive subspaces while enhancing semantic alignment across modalities.

    \item We introduce an uncertainty-aware fusion mechanism that leverages posterior variance as a reliability indicator to dynamically adjust representation weights, yielding robust multimodal representations under missingness.

    \item Comprehensive experiments on three public MSA benchmarks demonstrate that MIDAS establishes a new state-of-the-art performance in incomplete scenarios, and ablation studies confirm the necessity and effectiveness of each component.
\end{itemize}

The remainder of this paper is organized as follows: In Section II, we briefly review some recent works on multimodal sentiment analysis, mutual information, and uncertainty modeling and fusion. In Section III, we introduce the framework of our proposed method in detail. In Section IV, we illustrate the experiments and setups. In Section V, we present and analyze the experimental results. Finally, we summarize this paper and discuss future work in Section VI.

\section{Related Work}

\subsection{Multimodal Sentiment Analysis}

Recent studies \cite{yu2021Self-MM,han2021MMIM,wang2023TETFN,wen2025dashfuison,zhang2026improving} have primarily focused on developing advanced alignment and fusion mechanisms to improve MSA performance.
For instance, MISA \cite{hazarika2020MISA} projects each modality into modality-invariant and modality-specific spaces and then fuses them.
However, these methods typically assume complete inputs, which limits their applicability in real-world scenarios. To address this issue, several approaches have been proposed for incomplete MSA, which can be broadly categorized into two groups:

\textbf{Data imputation-based methods} aim to reconstruct missing modalities prior to downstream fusion. For instance, TFR-Net \cite{yuan2021tfr-net} employs a Transformer-based reconstruction module to infer missing features, while NIAT \cite{yuan2023NIAT} leverages adversarial learning \cite{goodfellow2020GAN} to obtain noise-invariant representations. CIF-MMIN \cite{liu2024contrastive} applies contrastive learning to extract modality-invariant features and reconstruct absent information. EMT \cite{sun2023efficient} proposes a dual-level restoration module that reconstructs low-level features and learns high-level semantics via siamese representation learning. IMDer \cite{wang2023incomplete} uses a score-based diffusion model conditioned on available modalities to recover missing data distributions.
While these methods yield promising results, they often introduce substantial computational overhead. Moreover, most rely on the presence of at least one completely available modality, which may not be guaranteed in real-world scenarios.

\textbf{Coordinated representation-based methods} focus on learning robust representations from incomplete multimodal inputs directly. CorrKD \cite{li2024correlation} transfers cross-sample knowledge to restore missing semantics through the contrastive distillation framework. LNLN \cite{zhang2024lnln} incorporates language-guided noise-resistant learning to enhance robustness, building on ALMT \cite{zhang2023ALMT}. UMDF \cite{li2024UMDF} introduces a self-distillation framework to learn robust representations from consistent multimodal distributions. GCNet \cite{lian2023gcnet} leverages two graph neural networks to capture temporal and speaker information in conversations.
Although these methods are more efficient and scalable, they typically operate directly in the original feature space, where semantic inconsistencies and distortions can limit the quality of the learned representations. 
Our method explicitly disentangles modality-shared and modality-exclusive components via a mutual information minimax framework, enabling more robust and semantically coherent representation learning under incomplete multimodal conditions.

\subsection{Mutual Information}

Mutual information (MI) is a fundamental concept in information theory that measures the relationship between two random variables. It is a reparameterization-invariant measure of dependency:
\begin{equation}
I(X;Y)=\mathbb{E}_{(X, Y) \sim p(x, y)} \left[ \log \frac{p(X, Y)}{p(X)p(Y)} \right].
\end{equation}
Further, the mutual information of three random variables is called interaction information, which is defined as:
\begin{equation}
\begin{aligned}
I(X;Y;Z) &= I(X;Y) - I(X;Y|Z) \\
&= I(X;Z) - I(X;Z|Y) \\
&= I(Y;Z) - I(Y;Z|X).
\end{aligned}
\end{equation}

Tishby et al. \cite{tishby2015deepIB} first used the Information Bottleneck principle to help understand neural networks, and variational information bottleneck \cite{alemi2016deepVIB} introduced MI-based regularization into deep models. Since then, MI maximization has been widely adopted in representation learning \cite{hjelm2018deepinfomax,bachman2019learning,tschannen2019mutual}, aiming to preserve relevant information in learned features.
Due to the difficulty of estimating MI in high-dimensional spaces, many works propose variational bounds or neural estimators such as MINE \cite{belghazi2018MINE}, InfoNCE \cite{oord2018representation}, and CLUB \cite{cheng2020club}.

In MSA, several studies have explored MI-based techniques to improve multimodal representation learning.
Han et al.\cite{han2021MMIM} maximized mutual information between modality features to improve fusion performance. \cite{zheng2022multimodal} further minimized the MI between input data and corresponding features to mine task-related information. 
However, these methods typically employ MI only for fusion, overlooking the capture of modality-specific details. More recently, \cite{sun2025multimodal} used MI and conditional MI to estimate modality-invariant, -specific, and -complementary information. Nonetheless, their approach relies on a complex combination of NMJ and MINE estimators, and does not explicitly disentangle modality features, which limits its effectiveness in handling incomplete inputs. 
In contrast, our MIDAS framework adopts a simple yet effective dual-MI objective: minimizing MI between shared and exclusive features for disentanglement, while maximizing MI across shared features for alignment. This design enables the model to learn well-separated and interpretable representations, improving robustness under incomplete input conditions.

\subsection{Uncertainty-aware Modeling and Fusion}

Uncertainty provides a principled way to quantify prediction confidence and assess model reliability \cite{gawlikowski2023survey}. In deep learning, uncertainty is typically categorized into epistemic uncertainty, which reflects uncertainty in model parameters, and aleatoric uncertainty, which captures noise inherent in the data \cite{kendall2017uncertainties}. A variety of approaches have been proposed to estimate uncertainty, including Bayesian neural networks \cite{denker1990transforming, neal2012bayesian}, Monte Carlo Dropout \cite{gal2016dropout}, deep ensembles \cite{lakshminarayanan2017simple}, predictive modeling \cite{kendall2017uncertainties}, and energy-based techniques \cite{liu2020energy}. Leveraging uncertainty has been shown to improve robustness across several domains, such as semantic segmentation \cite{yang2023uncertainty}, action understanding \cite{guo2024uncertainty}, and facial expression recognition \cite{zhang2021relative}.

In multimodal learning, uncertainty naturally emerges due to heterogeneous sensing mechanisms, modality-specific noise, and inconsistent data quality. Recent works have increasingly leveraged uncertainty to enhance multimodal fusion \cite{han2023TMC,chen2025uncertainty} as well as to improve performance on downstream tasks \cite{zhou2024uncertainty,shao2025ua}. Subedar et al. \cite{subedar2019uncertainty} applied Bayesian deep learning and accuracy-vs-uncertainty trade-offs to guide fusion decisions. EAU \cite{gao2024embracing} modeled modality-wise aleatoric uncertainty to stabilize joint representations under noisy multimodal inputs. COLD Fusion \cite{Tellamekala2024cold} further leveraged Gaussian uncertainty representations to quantify modality reliability and regulate fusion variance.
However, most existing multimodal uncertainty methods rely on auxiliary predictors or sampling-based estimates, which introduce additional model complexity and computational cost. In contrast, our method directly utilizes the intrinsic posterior uncertainty derived from variational modeling and incorporates it into an uncertainty-aware fusion mechanism, enabling a more reliable and robust fusion strategy under incomplete multimodal conditions.

\begin{figure*}[ht]
  \centering
  \includegraphics[width=0.97\textwidth]{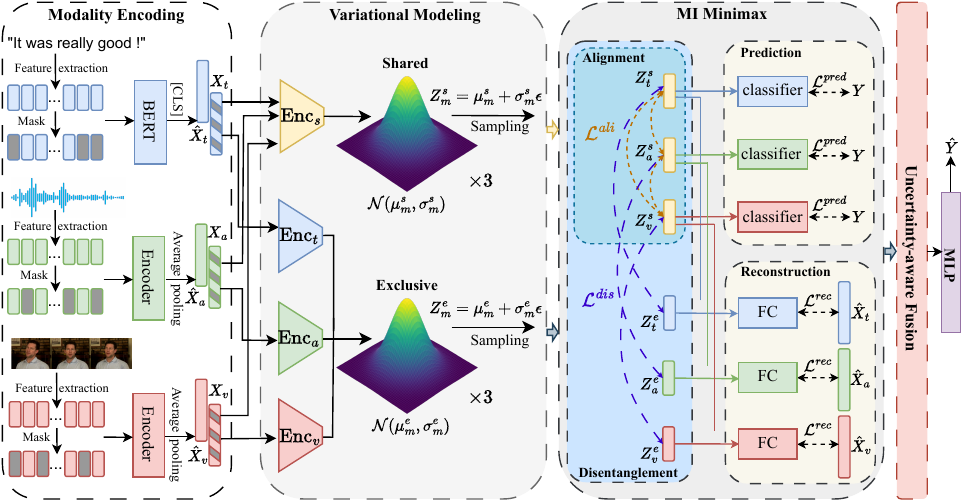}
  \caption{The overall architecture of the MIDAS for incomplete multimodal sentiment analysis. It consist of modality encocding, vatiational modeling, MI minimax, and uncertainity-aware fusion. Features extracted from raw inputs are first randomly masked and encoded independently, factorized into shared and exclusive latent distribution for disentanglement and alignment, finally fused via uncertainty-aware fusionto produce robust sentiment predictions.}
  \label{fig:framework}
\end{figure*}

\section{Method}

The overall architecture of the proposed MIDAS framework is shown in Fig.~\ref{fig:framework}. 
To effectively handle real-world incompleteness, we first construct multimodal inputs with stochastically missing data. 
Each raw modality input is then encoded into a feature sequence and passed through the variational modeling module, which factorizes each modality into shared and exclusive latent distributions for subsequent MI minimax.
Specifically, we minimize the mutual information between shared and exclusive latent spaces to achieve clean disentanglement, while maximizing the mutual information across shared spaces encourages cross-modal semantic alignment. To further stabilize learning and preserve meaningful semantic content, prediction and reconstruction objectives are incorporated to guide the disentangled representations. Finally, an uncertainty-aware fusion module adaptively integrates shared and exclusive representations by leveraging posterior uncertainty, yielding reliable sentiment predictions.

\subsection{Input Construction and Modality Encoding}

\subsubsection{Input Construction}
Our multimodal inputs consist of video, audio, and text data. These raw inputs are first transformed into feature sequences by widely-used feature extractors. After that, we construct incomplete data. For each modality, we randomly erase a varying proportion of the information. Specifically, for the visual and audio modalities, erased segments are replaced with zeros. For the text, we follow the recent work \cite{yuan2021tfr-net} to substitute erased tokens with [UNK]. Unlike [MASK] is strongly coupled with its Masked Language Modeling (MLM) pre-training objective for predicting missing words, [UNK] represents unknown tokens in the BERT vocabulary \cite{devlin2018bert}, which better simulate real-world noise.

\subsubsection{Modality Encoding}
For each sample, we encode both complete and incomplete modality sequences using shared modality-specific encoders to obtain consistent feature representations. For the visual and audio modalities, we employ two separate Transformer encoders to capture temporal dependencies and contextual relationships within each sequence. The encoder outputs are aggregated via average pooling to produce visual and audio features. For the text modality, we use a pre-trained BERT model to encode the input sequence, extracting the hidden state of the [CLS] token as the text feature. To ensure dimensional consistency across modalities, the BERT output is projected to match the feature dimensions of the audio and visual features.
We denote the resulting incomplete and complete modality features as $X_m \in \mathbb{R}^{d}$ and $\hat{X}_m \in \mathbb{R}^{d}$, respectively, where $m \in \{t, v, a\}$ corresponds to the text, visual, and audio modalities. Here, $d$ represents the feature dimension.

\subsection{Variational Modeling}

To capture the inherent uncertainty and variability caused by incomplete multimodal data, we adopt a variational inference strategy that models latent variables as multivariate Gaussian distributions instead of deterministic point embeddings \cite{wei2024DMRNet}. This probabilistic formulation decomposes each modality into a pair of latent distributions: one shared and one exclusive:

\begin{equation}
\begin{gathered}
p\left(Z_m^s \mid X_m\right) \sim \mathcal{N}\left(\mu_m^s, \sigma_m^{s} I \right), \\
p\left(Z_m^e \mid X_m\right) \sim \mathcal{N}\left(\mu_m^e, \sigma_m^{e} I \right),
\end{gathered}
\end{equation}
where $Z_m^{s}$ and $Z_m^{e}$ denote the shared and exclusive latent features, respectively, and $I$ is the identity matrix.
We utilize a shared encoder $\text{Enc}_s$ and three modality-exclusive encoders $\text{Enc}_e$ to model these distributions. Each encoder consists of fully connected layers that estimate the mean $\mu$ and variance $\sigma$. Following standard practice for numerical stability, we predict the log-variance $\log \sigma^2$ \cite{chun2021probabilistic}:
\begin{equation}
\mu_m^s, \sigma_m^s = Enc_s(X_m), \ \mu_m^e, \sigma_m^e = Enc_m(X_m),
\end{equation}
In this framework, the representation is treated as a stochastic sample from a distribution rather than a fixed embedding. Crucially, the variance $\sigma_m$ quantifies the aleatoric uncertainty associated with modality $m$, while the mean $\mu_m$ serves as the stable representation.

However, directly sampling from the latent distributions is non-differentiable. To enable end-to-end optimization, we apply the reparameterization trick \cite{kingma2013VAE} to sample latent variables:
\begin{equation}
Z_{m}^{s} = \mu_{m}^{s} + \sigma_{m}^{s} \odot \epsilon, \;
Z_{m}^{e} = \mu_{m}^{e} + \sigma_{m}^{e} \odot \epsilon, \; \epsilon \sim \mathcal{N}(0, I),
\end{equation}
where $\odot$ denotes element-wise multiplication. During training, we sample $Z_m^{s}$ and $Z_m^{e}$ for downstream processing, while at inference time we use the means $\mu_m^{s}$ and $\mu_m^{e}$ as deterministic embeddings for stable prediction.

\subsection{MI Minimax}

To enable flexible and robust multimodal representation learning, we disentangle each modality feature \( X_m \) into two distinct latent components: a shared representation that captures modality-invariant information and an exclusive representation that retains modality-exclusive characteristics.

\subsubsection{Disentanglement via MI Minimization}

To ensure that shared and exclusive representations capture distinct factors of variation, we introduce a regularization term that minimizes the mutual information between them.
\begin{equation}
\mathcal{L}_k^{\text{dis}}= \frac{1}{3}  \sum_{m\in\{t,a,v\}} I\left( Z_m^{s} ; Z_m^{e} \right),
\end{equation}
where \(\mathcal{L}_k^{\text{dis}} \) denotes the disentanglement loss of the $k$-th sample. This encourages statistical independence between shared and exclusive representations, promoting effective disentanglement.

However, computing mutual information directly between hidden representations is intractable. Inspired by the strategy proposed in \cite{Zhang2024FedDCSR}, we introduce the input \( X_m \) as an auxiliary variable. Leveraging the interaction information, the MI between $Z_m^{s}$ and $Z_m^{e}$ can be reformulated as:

\begin{equation}
\begin{aligned}
I(Z_m^{s} ; Z_m^{e}) =& I(Z_m^{s};X_m) - I(Z_m^{s};X_m | Z_m^{e}) \\
&+ I(Z_m^{s} ; Z_m^{e} | X_m).
\end{aligned}
\end{equation}

Under the assumption that conditioning on $X_m$ renders $Z_m^{s}$ and $Z_m^{e}$ statistically independent of any additional side-information, the posterior distribution $q(Z_m^{s} | X_m) = q(Z_m^{s} | X_m, Z_m^{e})$. Consequently, the conditional mutual information satisfies \(I(Z_m^{s} ; Z_m^{e} | X_m)=0\), allowing the third term in Eq. (7) to be removed.

According to the chain rule of mutual information, $I(X_m;Z_m^s,Z_m^e)=I(X_m;Z_m^e)+I(X_m;Z_m^s | Z_m^e)$. Leveraging the symmetry of mutual information, we can rearrange this to yield $I({Z_m^s;X}_m| Z_m^e)=-I(Z_m^e;X_m)+I(X_m;Z_m^s,Z_m^e)$.
Thus, $I(Z_m^{s};Z_m^{e})$ can be expanded and written as:
\begin{equation}
I(Z_m^{s};Z_m^{e}) = I(Z_m^{s};X_m)+I(Z_m^{e};X_m)-I(X_m;Z_m^{s},Z_m^{e}).
\end{equation}

Based on this decomposition, each term can be upper-bounded using variational approximations. Through introducing variational encoders $q(Z^{s}_m|X_m)$, $q(Z^{e}_m|X_m)$, and a variational decoder $p(X_m \mid Z^{s}_m, Z^{e}_m)$, we derive a tractable variational upper bound using KL divergence and reconstruction likelihood \cite{hwang2020variational}. Consequently, we optimize the following variational upper bound.
\begin{align}
I(Z^{s}_m ; Z^{e}_m) 
&\leq \mathbb{E}_{p(X_m)} \Big[
  D_{\mathrm{KL}}\big( q(Z^{s}_m|X_m) \,\|\, p(Z^{s}_m) \big) \notag\\
&\quad + D_{\mathrm{KL}}\big( q(Z^{e}_m|X_m) \,\|\, p(Z^{e}_m) \big) \Big] \notag\\
&\quad - \mathbb{E}_{p(Z^{s}_m, Z^{e}_m)} 
    \mathbb{E}_{q(Z^{s}_m|X_m) q(Z^{e}_m|X_m)} \notag\\
&\quad\quad \Big[ \log p(X_m \mid Z^{s}_m, Z^{e}_m) \Big].
\end{align}

Here, \( p(X_m) \) is typically chosen as a standard Gaussian prior, and \( p(X_m|Z_m^{s},Z_m^{e}) \) denotes the reconstruction likelihood of \( X_m \) of the input given its latent representations \( (Z_m^{s},Z_m^{e}) \).
This formulation not only encourages independence between the shared and exclusive latent spaces but also ensures that both remain informative and coherent with respect to the original modality.

\subsubsection{Alignment via MI Maximization}

To ensure that the shared latent representations capture modality-invariant semantics, we explicitly encourage alignment across modalities by maximizing the mutual information between shared representations from different modalities. This promotes a common latent structure that is robust to modality-exclusive noise and facilitates cross-modal understanding.
Formally, let \( \mathcal{P} = \bigl\{(t,a),\,(t,v),\,(a,v)\bigr\} \)
be the set of all unordered modality pairs. The alignment loss of the $k$-th sample is defined as the (negative) average mutual information across these pairs:
\begin{equation}
\mathcal{L}_k^{\mathrm{ali}}
= -\frac{1}{|\mathcal{P}|}\sum_{(m_1,m_2)\in\mathcal{P}}
I\bigl(Z^{s}_{m_1};Z^{s}_{m_2}\bigr).
\end{equation}

Since exact mutual information estimation between latent variables is intractable, we adopt the Jensen–Shannon Divergence (JSD) estimator from Deep InfoMax \cite{hjelm2018deepinfomax}, which provides a stable and sample-efficient variational lower bound. Compared with InfoNCE, which requires a large number of negative samples, the JSD estimator is more robust and better suited to our setting. Let \( f_\phi( \cdot;\cdot)\) be a critic network that distinguishes positive (joint) pairs from negative (marginal) ones. The variational lower bound is written as:
\begin{equation}
\begin{aligned}
I\bigl(Z^{s}_{m_1};Z^{s}_{m_2}\bigr) &\ge
\mathrm{JSD}\bigl(q(Z^{s}_{m_1},Z^{s}_{m_2})\;\|\;q(Z^{s}_{m_1})\,q(Z^{s}_{m_2})\bigr) \\
&=\mathbb{E}_{q(Z^{s}_{m_1},Z^{s}_{m_2})}\bigl[-\mathrm{sp}\bigl(-f_\phi(Z^{s}_{m_1},Z^{s}_{m_2})\bigr)\bigr] \\ 
&-\mathbb{E}_{q(Z^{s}_{m_1})q(Z^{s}_{m_2})}\bigl[\mathrm{sp}\bigl(f_\phi(Z^{s}_{m_1},{Z^{s}_{m_2}}')\bigr)\bigr],
\end{aligned}
\end{equation}
where \(\mathrm{sp}(u)=\log(1+e^u)\) is the softplus function, and \({Z^{s}_{m_2}}'\) denotes a negative sample drawn from the marginal distribution. The critic \( f_\phi \) is implemented as a simple MLP.
To strengthen the discriminative ability of the critic, we construct hard negative samples by replacing \( {Z^{s}_{m_2}}' \) with exclusive features \( Z^{e}_{m_2} \). This forces the critic to distinguish shared semantics from modality-exclusive information, thereby strengthening the alignment objective.

The first expectation term encourages the critic to assign high confidence to true joint samples, while the second penalizes it for assigning high confidence to unrelated (marginal) pairs. Maximizing this JSD-based lower bound effectively brings shared latent distributions across modalities closer together, promoting consistent and semantically aligned representations.
This MI-based alignment is particularly beneficial in settings where one or more modalities are incomplete or unreliable, as it ensures that shared features remain informative and semantically aligned across views.

\subsubsection{Prediction}

Although MI maximization encourages shared representations to be aligned across modalities, it does not inherently ensure that these shared features are semantically meaningful for the downstream sentiment task. This limitation becomes more pronounced when the multimodal inputs are incomplete, where modality features inevitably contain a mixture of semantic content and noise. Without additional constraints, the shared latent space may collapse to trivial commonalities or noise-corrupted patterns.
To prevent this and to anchor the shared representations to task-relevant semantics, we introduce an auxiliary prediction objective. Since each individual shared representation \( Z_m^{s} \) may not contain sufficient information to predict the full sentiment score, we adopt a lightweight binary classification task as weak supervision. For each modality, a separate linear classifier is applied to its shared representation, and the predictions are supervised by the ground-truth sentiment label \( y \) using cross-entropy loss:
\begin{equation}
\mathcal{L}^{\text{aux}}_k = \frac{1}{3} \sum_{m\in\{t,a,v\}} \mathcal{L}_{\text{CE}}\left( \text{classifier}_m(Z_m^{s}), y_k^{bi} \right),
\end{equation}
where \( \text{classifier}_m(\cdot) \) denotes the modality-exclusive classifier and \( \mathcal{L}_{\text{CE}} \) is the cross-entropy loss. This weak supervision explicitly encourages the shared encoders to extract discriminative features aligned with the target semantics, ensuring that \( Z_m^{s} \) becomes more reliable and informative for subsequent fusion.

\subsubsection{Reconstruction}

To ensure that the disentangled features retain enough information for the downstream task, we introduce a reconstruction loss. Rather than reconstructing the potentially incomplete features $X_m$, we use the complete features $\hat{X}_m$ as supervision. This strategy not only prevents trivial or degenerate solutions, where the encoder learns orthogonal yet uninformative representations, but also encourages the model to infer missing content and capture richer details.

For each modality, we concatenate the shared representation $Z_m^{s}$ and the exclusive representation $Z_m^{e}$, and pass the combined vector through a fully connected layer (FC) to project it back to the original input space. We use the mean squared error (MSE) between the reconstructed and original complete features as the reconstruction loss:
\begin{equation}
\mathcal{L}_k^{\text{rec}} = \frac{1}{3} \sum_{m\in\{t,a,v\}}\left\| \text{FC}(\text{concat}(Z_m^{s}; Z_m^{e})) - \hat{X}_m \right\|^2 ,
\end{equation}
where $\text{concat}(\cdot;\cdot)$ denotes concatenation. This loss ensures that both shared and exclusive features contribute meaningfully to reconstructing modality content.

\begin{figure*}[ht]
  \centering
  \includegraphics[width=0.93\textwidth]{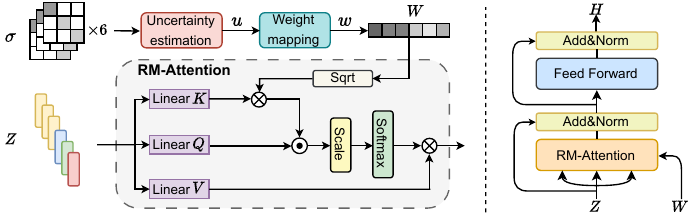}
  \caption{The framework of the proposed Uncertainty-aware Fusion. Posterior uncertainty estimates derived from latent variances are mapped to reliability weights W. These weights element-wise modulate the key matrix (K) within the Reliability-Modulated Attention (RM-Attn) block to emphasize reliable tokens during cross-modal fusion, producing the final fused representation H.}
  \label{fig:UAF}
\end{figure*}

\subsection{Uncertainty-aware Fusion}

Incomplete or corrupted modalities inevitably introduce uncertainty in the latent posterior distributions. However, most multimodal fusion modules operate under the assumption of equally reliable latent tokens, regardless of whether a modality is incomplete, highly noisy, or poorly reconstructed. This design may cause unreliable modalities to dominate cross-modal interactions, while informative ones are under-utilized.
To mitigate these issues, we introduce an uncertainty-aware fusion module that explicitly incorporates posterior uncertainty as a reliability indicator during cross-modal fusion. 

The structure of the uncertainty-aware fusion is illustrated in Fig.~\ref{fig:UAF}. For each modality $m$, we obtain two Gaussian posteriors from the variational modeling: the shared posterior and the exclusive posterior. For each sample, the shared feature $Z_m^{s}$ and exclusive feature $Z_m^{e}$ from modality $m$ are stacked into a unified token sequence:
\begin{equation}
Z=[Z_t^{s}, Z_a^{s},Z_v^{s}, Z_t^{e},Z_a^{e}, Z_v^{e}] \in \mathbb{R}^{6\times d}.
\end{equation}
We index these tokens by $j \in \{ 1,...,6\}$ and denote the per-token by $Z_j$, the per-token variances by $\sigma_j^2$.

\subsubsection{Posterior Uncertainty Estimation}

Given the variational posterior of modality $m$, the variance $\sigma_m$ reflects aleatoric uncertainty.
We estimate the uncertainty using the differential entropy of a diagonal Gaussian, without requiring additional estimators:
\begin{equation}
u_j=\frac{1}{2d} \sum_{i=1}^{d} \log\left(2\pi e \sigma_{j,i}^2\right).
\end{equation}
This entropy-based measure is scale-invariant and robust across encoders with different variance magnitudes.

\subsubsection{Mapping Uncertainty to Reliability Weights}

To integrate uncertainty into fusion, we map $u_m$ to a reliability weight $w_m\in(0,1)$.  
Following the logic of the provided implementation, we center the uncertainty values and apply a temperature-controlled sigmoid:

\begin{equation}
w_j = \sigma (-\tau (u_m - \frac{1}{6}\sum_{j=1}^6 u_j )), \qquad \tau>0,
\end{equation}
where $\sigma(\cdot)$ denotes the sigmoid function. This formulation suppresses tokens whose uncertainty is above the sample-wise mean while amplifying more reliable ones.

\subsubsection{Reliability-Modulated Attention}

We incorporate the reliability weights into the attention mechanism to prevent unreliable tokens from dominating cross-modal similarity. Let
\[
Q = ZW_Q,\quad K = ZW_K,\quad V = ZW_V,
\] 
where $W_Q$, $W_K \in \mathbb{R}^{d \times d_k}$, $W_V \in \mathbb{R}^{d \times d_v}$ are learnable parameters and $d_k$ is the dimension of the attention.
We form a diagonal reliability scaling matrix

The reliability-modulated attention (RM-Attention) is defined as:
\begin{equation}
\mathrm{RM\text{-}Attn}(Z)
= \mathrm{softmax}\left(
\frac{Q(\sqrt{W}K)^{\top}}{\sqrt{d}} \right)V,
\end{equation}

where $W = [w_1,...,w_6]\in\mathbb{R}^6$, which is expanded across the key dimension when applied inside attention. $\odot$ denotes element-wise modulation. Here, applying the raw weight directly may squash the logits of unreliable modalities too aggressively, forcing the softmax output into an uninformative uniform distribution. Therefore, we use $\sqrt{W}$ instead of $W$ for a softer temperature calibration to avoid overly shrinking. This ensures that while unreliable modalities are correctly attenuated and prevented from dominating cross-modal interactions, the variance of the logits is preserved enough to maintain healthy gradients and numerical stability during softmax computation.

As in standard Transformer blocks, the RM-Attention output is followed by residual connections, layer normalization, and a feed-forward network:
\begin{align}
H &= \mathrm{LayerNorm}\left(Z + \mathrm{RM\text{-}Attn}(Z)\right),\\
H &= \mathrm{LayerNorm}\left(H + \mathrm{FFN}(H)\right).
\end{align}

The resulting representation $H$ is then passed to downstream prediction layers. By explicitly grounding fusion in posterior uncertainty, this module adaptively emphasizes reliable latent tokens while suppressing noisy ones, improving robustness under incomplete and corrupted multimodal conditions.

\subsection{Overall Learning Objective}
After uncertainty-aware fusion, we obtain contextualized token representations $H\in \mathbb{R}^{6\times d}$. We flatten these tokens by concatenation to form a compact fused feature for downstream prediction:
\begin{equation}
H_{\text{fused}} = \text{concat}(H_1, H_2, \dots, H_6) \in \mathbb{R}^{6d},
\end{equation}
where \( H_i \) denotes the \(i\)-th token of \( H \) and \(H_{\text{fused}} \) is the fused feature.

The fused feature \(H_{\text{fused}}\) is passed through a task-specific decoder implemented as a MLP to generate the final prediction.
The task loss for the \(k\)-th training sample is defined by the MSE loss between the ground-truth label \( y_k \) and the model’s prediction \( \hat{y}_k \):
\begin{equation}
\mathcal{L}_k^{\text{task}} = \left\| y_k - \hat{y}_k \right\|^2.
\end{equation}

In summary, our method jointly optimizes four learning objectives to enable robust and interpretable multimodal representation learning.
(1) a \textbf{task loss} \( \mathcal{L}^{\text{task}} \) that guides the model toward accurate prediction;  
(2) a \textbf{disentanglement loss} \( \mathcal{L}^{\text{dis}} \) that  encourages independence between shared and exclusive features;  
(3) an \textbf{alignment loss} \( \mathcal{L}^{\text{ali}} \) that improves consistency among shared representations across modalities;
(4) a \textbf{prediction loss} \( \mathcal{L}^{\text{pred}} \) that anchors shared representations to task-relevant semantics; and
(5) a \textbf{reconstruction loss} \( \mathcal{L}^{\text{rec}} \) that ensures the retention of modality-exclusive information.
The overall training objective aggregates these losses over the training set of size $N$:
\begin{equation}
\mathcal{L} = \frac{1}{N} \sum_{k=1}^{N} \left( \mathcal{L}_k^{\text{task}} + \alpha \mathcal{L}_k^{\text{dis}} + \beta \mathcal{L}_k^{\text{ali}} + \gamma \mathcal{L}_k^{\text{pred}} + \lambda  \mathcal{L}_k^{\text{rec}}  \right),
\end{equation}
where $ \alpha $, $ \beta $, $ \gamma$, and $\lambda$ are scalar hyperparameters that balance the relative importance of auxiliary objectives. All components are optimized jointly by gradient descent.

\section{Experimental Databases and Setup}

In this section, we first describe three benchmark conversational datasets in our experiments. Following that, we illustrate
evaluation metrics, baselines, and multimodal features in detail. Finally, we introduce implementation details, including experiments, models, and training details.

\subsection{Datasets}

To evaluate the performance of MIDAS, we conduct experiments on three publicly available datasets in MSA research, MOSI \cite{zadeh2016mosi}, MOSEI \cite{zadeh2018mosei}, and CH-SIMS \cite{yu2020ch-sims}.

\textbf{MOSI.}
The MOSI \cite{zadeh2016mosi} is a widely used benchmark dataset for MSA, containing 2,199 utterance-level video clips segmented from 93 opinion videos. The dataset is split into 1,284 training samples, 229 validation samples, and 686 test samples. Each clip is manually annotated with a sentiment score ranging from -3 (strongly negative) to +3 (strongly positive).

\textbf{MOSEI.}
The MOSEI \cite{zadeh2018mosei} extends MOSI in both size and diversity, comprising 22,856 labeled video clips collected from YouTube. It is divided into 16,326 for training, 1,871 for validation, and 4,659 for testing. The annotation scheme is consistent with MOSI.

\textbf{CH-SIMS.}
The CH-SIMS \cite{yu2020ch-sims} is a Chinese MSA dataset containing 2,281 selected video clips. The dataset is split into 1,368 training, 456 validation, and 457 test samples. Each sample is annotated with one multimodal sentiment label and three unimodal labels, with sentiment scored from -1 (strongly negative) to +1 (strongly positive).

\subsection{Baselines and Evaluation Metrics}

\textbf{Baselines.}
To thoroughly assess the effectiveness of our proposed method, we compare it against several strong and state-of-the-art MSA models, including complete MSA methods: MISA \cite{hazarika2020MISA}, Self-MM \cite{yu2021Self-MM}, MMIM \cite{han2021MMIM}, CENET \cite{wang2023CENET}, TETFN \cite{wang2023TETFN}, ALMT \cite{zhang2023ALMT}; as well as incomplete MSA methods: TFR-Net \cite{yuan2021tfr-net}, EMT-DLFR \cite{sun2023efficient}, LNLN \cite{zhang2024lnln}, and P-RMF \cite{zhu2025proxy}. To ensure fairness in comparison, methods that only report results from a single run and lack valid official code for reproduction are not selected.

\textbf{Evaluation Metrics.}
Following previous works \cite{zhang2024lnln}, we evaluate both classification and regression performance for a comprehensive comparison with baselines. For classification, we report binary (Acc-2), ternary (Acc-3), 5-class (Acc-5), and 7-class (Acc-7) accuracy, as well as weighted F1 scores. For binary accuracy, we adopt two settings: (1) Negative/Positive (excluding zero from both classes), and (2) Negative/Non-negative (including zero in the non-negative class). For regression, we report Mean Absolute Error (MAE) and the Pearson correlation coefficient (Corr). Except for MAE, higher values indicate better performance for all metrics.

\begin{table}[t]
\centering
\small
\setlength{\tabcolsep}{6pt}  
\caption{Hyperparameter configurations for MIDAS on MOSI, MOSEI, and CH-SIMS datasets}
\label{tab:hyperparams}
\begin{tabular}{lccc}
\toprule
\textbf{Hyperparameter} & \textbf{ MOSI} & \textbf{ MOSEI} & \textbf{CH-SIMS} \\
\midrule
Vector Dimension $d$          & 128   & 128   & 128 \\
Encoder Layers                & 2     & 2     & 2 \\
Batch Size                    & 32    & 32    & 32 \\
Training Epochs               & 100   & 20    & 100 \\
Initial LR (BERT)             & 1e-5  & 1e-5  & 1e-5 \\
Initial LR (others)           & 2e-3  & 1e-3  & 1e-3 \\
Weight Decay (BERT)           & 1e-4  & 1e-4  & 1e-4 \\
Weight Decay (others)         & 1e-3  & 1e-3  & 1e-3 \\
Loss Weights $\alpha$         & 0.05  & 0.05  & 0.05 \\
Loss Weights $\beta$          & 1.0   & 1.0   & 1.0 \\
Loss Weights $\gamma$         & 0.1   & 0.1   & 0.1 \\
Loss Weights $\lambda$        & 1.0   & 1.0   & 1.0 \\
Temperature $\tau$            & 10    & 10    & 10 \\
\bottomrule
\end{tabular}
\end{table}

\subsection{Feature Extraction}

To ensure a fair comparison, we leverage the official unaligned features provided by each benchmark dataset, consistent with the top-performing MSA methods.

\textbf{Text Modality}: Transformer-based pre-trained language models have achieved state-of-the-art results across various natural language processing (NLP) tasks. Following recent works \cite{zhang2024lnln}, we use pre-trained BERT models to encode raw text. Specifically,  "bert-base-uncased" \footnote{\url{https://huggingface.co/bert-base-uncased/}} is used for MOSI and MOSEI, while "bert-base-chinese" \footnote{\url{https://huggingface.co/bert-base-chinese/}} is employed for CH-SIMS.

\textbf{Audio Modality}: For MOSI and MOSEI, we use the COVAREP framework \cite{2014COVAREP} to extract low-level audio features, including pitch, voiced/unvoiced segments, glottal source parameters, and 12 Mel-frequency cepstral coefficients (MFCCs), among others. For CH-SIMS, we employ Librosa \cite{mcfee2015librosa} to obtain audio representations, extracting features such as the logarithmic fundamental frequency (log F0), 20 MFCCs, and 12 Constant-Q Transform (CQT) chromagrams.

\textbf{Vision Modality}: For MOSI and MOSEI, Facet \footnote{\url{https://imotions.com/platform/}} is utilized to extract 35 facial action units, capturing facial muscle movements linked to emotions. For CH-SIMS, we use the OpenFace2.0 toolkit \cite{OpenFace} to obtain visual features, including facial action units, facial landmarks, and head pose. These visual features are sampled at 30 Hz, forming a sequence that captures facial gestures over time.

\begin{table*}[t]
\caption{Robustness comparisons of the overall performance on MOSI and MOSEI datasets. In Acc-2 and F1, the left of the "/" corresponds to "Negative/Positive" and the right corresponds to "Negative/Non-negative". Best results are \textbf{bolded}, second results are \underline{underlined}. Note: Higher metrics denote better performance, except for MAE where lower is better.}
  \centering
  \renewcommand{\arraystretch}{1.}
  \small  
  \resizebox{\textwidth}{!}{%
  \begin{tabular}{lcccccccccccc}
    \toprule
    \multirow{2}{*}{Method}
      & \multicolumn{6}{c}{MOSI}
      & \multicolumn{6}{c}{MOSEI} \\
    \cmidrule(lr){2-7} \cmidrule(lr){8-13}
    & Acc-2 & F1 & Acc-5 & Acc-7 & MAE & Corr
    & Acc-2 & F1 & Acc-5 & Acc-7 & MAE & Corr \\
    \midrule
    MISA\cite{hazarika2020MISA}     & 70.89/68.87 & 67.37/64.89 & 32.95 & 30.25 & 1.100 & 0.510 
             & 72.73/76.44 & 66.82/69.95 & 38.70 & 37.90 & 0.791 & 0.497 \\
    Self-MM\cite{yu2021Self-MM}  & 70.87/69.30 & 68.74/66.75 & 33.65 & 30.82 & \textbf{1.065} & \underline{0.522}
             & 73.66/76.60 & 69.73/73.53 & 46.46 & 45.78 & 0.701 & 0.496 \\
    MMIM\cite{han2021MMIM}     & 69.26/68.23 & 67.94/66.72 & 34.56 & 31.43 & 1.108 & 0.485
             & 74.55/76.83 & 71.10/74.21 & 44.75 & 43.89 & 0.725 & 0.501 \\
    CENET\cite{wang2023CENET}    & 70.43/68.19 & 66.79/64.01 & 33.55 & 30.54 & \underline{1.069} & 0.511
             & 73.85/\underline{76.94} & 68.84/72.93 & 45.70 & 44.88 & 0.719 & 0.531 \\
    TETFN\cite{wang2023TETFN}    & 71.03/69.42 & 69.69/67.67 & 33.85 & 30.36 & \textbf{1.065} & 0.518
             & 74.90/76.91 & 71.50/74.34 & 44.72 & 43.60 & 0.720 & 0.516 \\
    ALMT\cite{zhang2023ALMT}     & 69.97/69.32 & 69.94/69.21 & 34.68 & \underline{31.49} & 1.128 & 0.493
             & 74.91/76.47 & 71.64/74.07 & 45.66 & 45.17 & 0.707 & 0.511 \\
    \midrule
    TFR-Net\cite{yuan2021tfr-net}  & \underline{71.16}/69.70 & 69.38/67.58 & 34.53 & 30.89 & \textbf{1.065} & 0.519 
             & 73.24/76.49 & 68.92/72.96 & 41.98 & 41.71 & 0.752 & 0.519 \\
    EMT-DLFR\cite{sun2023efficient}  & 71.04/\underline{70.26} & \underline{70.87}/\underline{70.18} & \underline{35.19} & 31.48 & 1.098 & 0.514 
             & 77.76/76.09 & 77.34/\underline{76.01} & \underline{48.23} & \underline{47.34} & 0.665 & 0.590 \\
    LNLN\cite{zhang2024lnln}     & 70.09/69.50 & 70.01/69.49 & 34.28 & 31.20 & 1.124 & 0.501 
             & 76.78/74.90 & 76.04/74.88 & 46.61 & 45.84 & 0.678 & 0.581 \\
    P-RMF\cite{zhu2025proxy}    & 69.92/69.10 & 69.82/68.53 & 33.42 & 31.03 & 1.134 & 0.501 
             & \underline{77.84}/74.68 & \underline{77.56}/75.36 & 47.96 & 46.94 & \underline{0.661} & \underline{0.595} \\
    \midrule
    \textbf{MIDAS}     & \textbf{71.88}/\textbf{71.26} & \textbf{71.91}/\textbf{71.20} & \textbf{35.12} & \textbf{32.06} &  1.074 & \textbf{0.534}&  \textbf{78.41}/\textbf{77.35} & \textbf{77.72}/\textbf{77.29} & \textbf{49.00} &\textbf{47.86} & \textbf{0.653} & \textbf{0.607} \\
  \bottomrule
  \end{tabular}
  }
  \label{robust_comparison:mosi}
\end{table*}

\begin{table}[t]
\caption{Robustness comparisons of the overall performance on CH-SIMS dataset. Best results are \textbf{bolded}, second results are \underline{underlined}. Note: Higher metrics denote better performance, except for MAE where lower is better.}
  \centering
  \small  
  \renewcommand{\arraystretch}{1.}  
  \resizebox{\columnwidth}{!}{%
  \begin{tabular}{lcccccc}
    \toprule
    Method & Acc-2 & F1 & Acc-3 & Acc-5 & MAE & Corr \\
    \midrule
    MISA\cite{hazarika2020MISA}     & 64.79 & 63.20 & 51.45 & 30.03 & 0.533 & 0.368 \\
    Self-MM\cite{yu2021Self-MM}  & 69.40 & 68.60 & 53.18 & 31.44 & \underline{0.507} & 0.391 \\
    MMIM\cite{han2021MMIM}     & 66.99 & 64.97 & 50.95 & 27.81 & 0.552 & 0.306 \\
    CENET\cite{wang2023CENET}    & 61.76 & 57.33 & 48.54 & 22.44 & 0.605 & 0.172 \\
    TETFN\cite{wang2023TETFN}    & 68.34 & 67.78 & 51.16 & 31.26 & 0.510 & 0.386 \\
    ALMT\cite{zhang2023ALMT}     & 68.81 & 68.14 & 53.02 & 32.29 & 0.542 & 0.326 \\
    \midrule
    TFR-Net\cite{yuan2021tfr-net}  & 63.27 & 59.22 & 49.96 & 26.98 & 0.593 & 0.192 \\
    EMT-DLFR\cite{sun2023efficient}  & \underline{71.41} & \underline{71.09} & \underline{57.66} & \underline{34.53} & 0.517 & \underline{0.403} \\
    LNLN\cite{zhang2024lnln}     & 69.43 & 68.04 & 54.33 & 31.24 & 0.540 & 0.323 \\
    P-RMF\cite{zhu2025proxy}    & 71.31 & 68.76 & 53.44 & 30.56 & 0.525 & 0.377 \\
    \midrule
    \textbf{MIDAS} & \textbf{73.68} & \textbf{72.15} & \textbf{57.72} & \textbf{35.08}  & \textbf{0.501} & \textbf{0.427} \\
    \bottomrule
  \end{tabular}
  }
  \label{robust_comparison:sims}
\end{table}

\subsection{Implementation Details}

\subsubsection{Experimental Details}

To thoroughly evaluate model robustness under real-world incomplete multimodal conditions, we apply random feature masking to each modality with predefined missing rates $r \in \{0.0,0.1,…,0.9\}$. Following the previous works \cite{yuan2021tfr-net,zhang2024lnln}, the same missing rate $r$ is uniformly applied across all modalities. For instance, $r=0.5$ indicates that 50\% of the features in the text, audio, and visual modalities are randomly removed.
Unlike prior works that train separate models for each missing rate, we train a single unified model capable of handling a wide range of missingness. Specifically, we split the training set into two equal parts: one half remains complete, while the other half is assigned a missing rate sampled uniformly from [0, 1].
Following standard practice, we adopt the official train/validation/test splits provided by the original works. Model selection is performed on the validation set under a fixed missing rate of 0.5. To ensure a realistic and fair evaluation, a single unified checkpoint is selected for each model, and its final performance across all metrics is reported on the test set under all missing rate conditions.

For baselines, the results of MISA, Self-MM, MMIM, CENET, TETFN, and TFR-Net are reproduced using the MMSA framework\footnote{\url{https://github.com/thuiar/MMSA}} \cite{mao2022M-SENA}.
The results of ALMT, LNLN, EMT-DLFR and P-RMF are reproduced using their official implementation. All baselines are trained with their recommended hyperparameters under the same experimental settings as our method.

\subsubsection{Model \& Training Details}

We set the latent feature dimension to $d=128$ for all datasets. Multimodal encoders consist of two single-layer Transformer encoders (128-dimensional) for audio and vision, respectively. For text, we directly use the pre-trained BERT encoder (768-dimensional) followed by a linear projection to 128 dimensions.
All models are implemented in PyTorch and trained on a single NVIDIA RTX 4090 GPU. We train MIDAS using the Adam optimizer \cite{kingma2014adam}. For SIMS and MOSI, we train the model for 100 epochs with a batch size of 32; for MOSEI, we train for 20 epochs with the same batch size. We use the warm-up strategy and cosine annealing scheduler for stable training.
Hyperparameter configurations for all datasets are summarized in Table~\ref{tab:hyperparams}.

\section{Results and Discussion}

\subsection{Robustness Comparison}

\begin{figure*}[t]
\newcommand{\imgwidth}{0.24\textwidth}
\newcommand{\rowtitle}[1]{%
  \raisebox{4.3em}{
    \begin{minipage}[c][4em][c]{1.5em}
      \centering
      \rotatebox{90}{\scriptsize\textbf{#1}}
    \end{minipage}%
  }%
}

\centering
\small

\makebox[\textwidth][l]{%
\begin{tabular}{@{}l@{\hspace{-0.5em}}c@{\hspace{0.4em}}c@{\hspace{0.6em}}c@{\hspace{0.4em}}c@{\hspace{-0.5em}}}

  & \scriptsize{Acc-2} & \scriptsize{F1} & \scriptsize{Acc-5} & \scriptsize{MAE} \\

  \rowtitle{MOSI} &
  \includegraphics[width=\imgwidth]{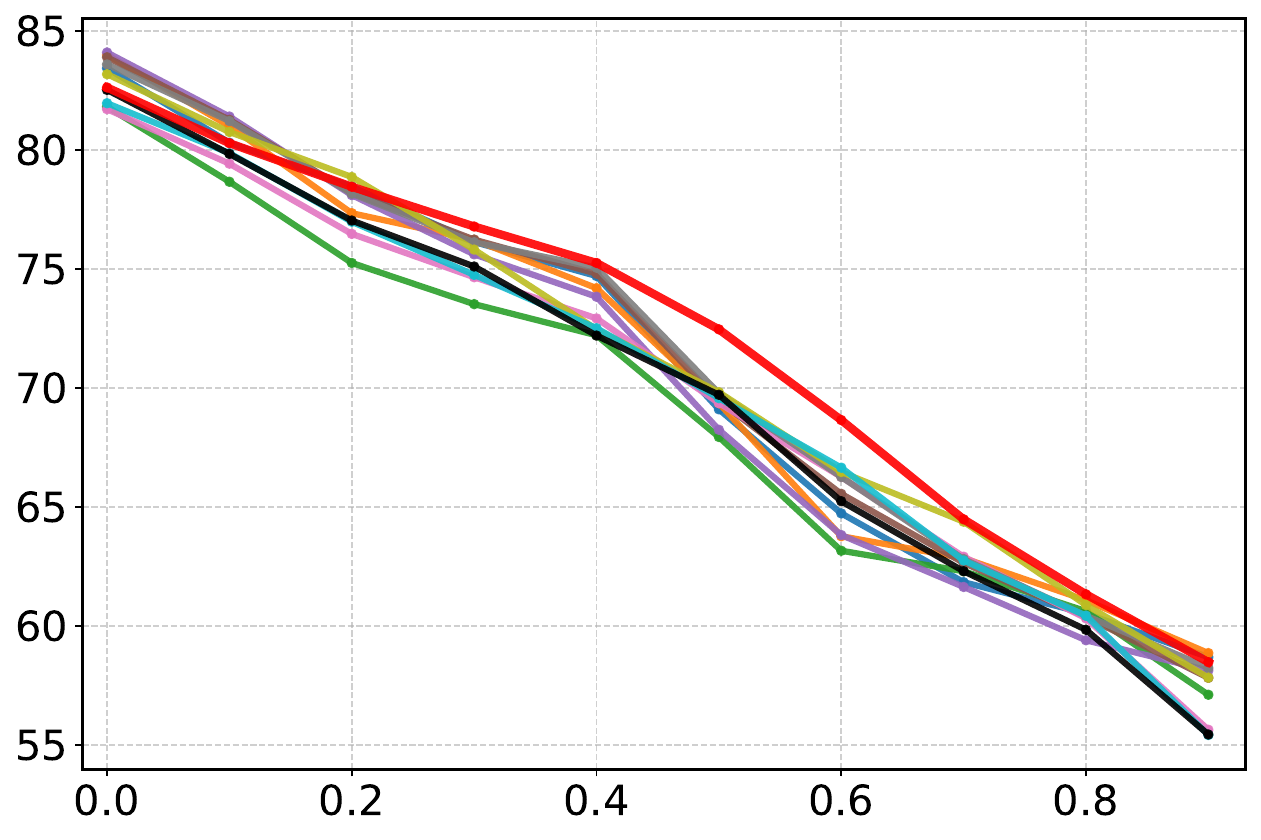} &
  \includegraphics[width=\imgwidth]{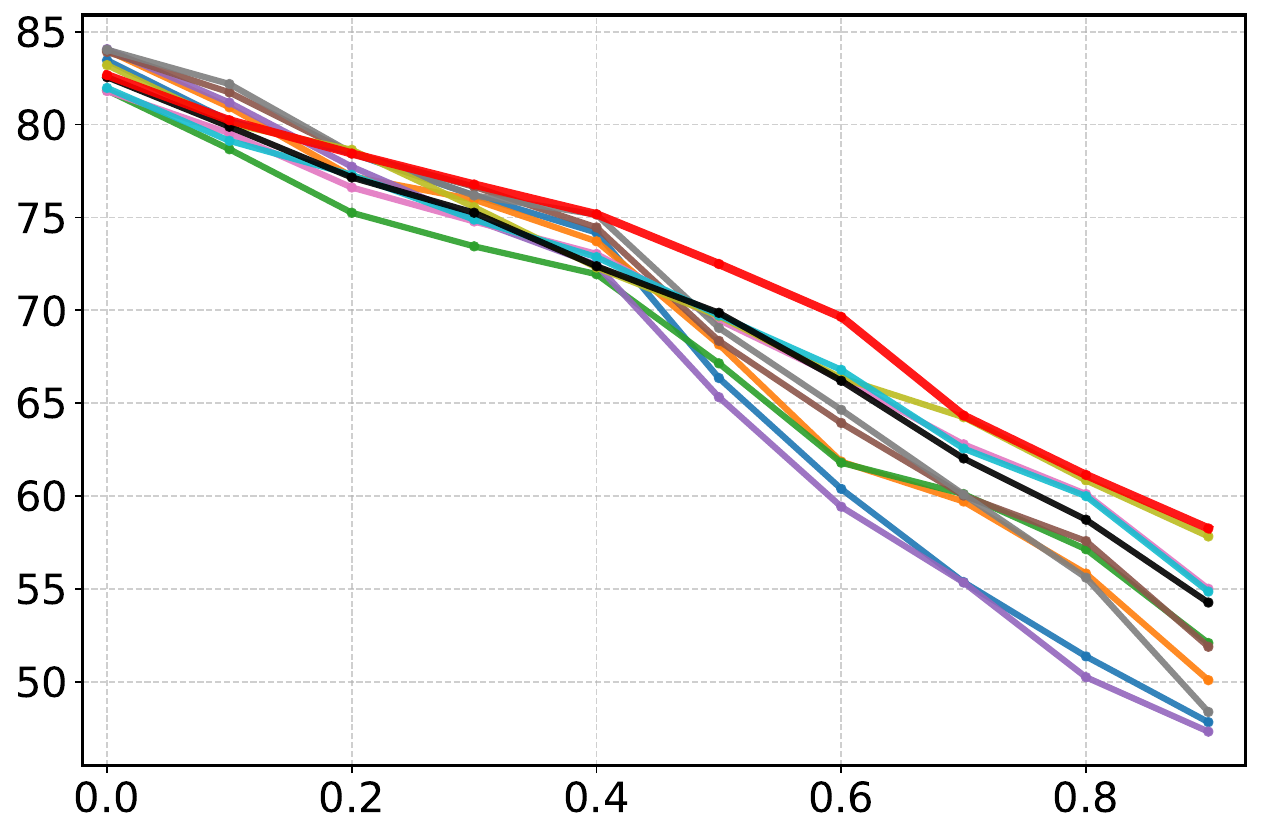} &
  \includegraphics[width=\imgwidth]{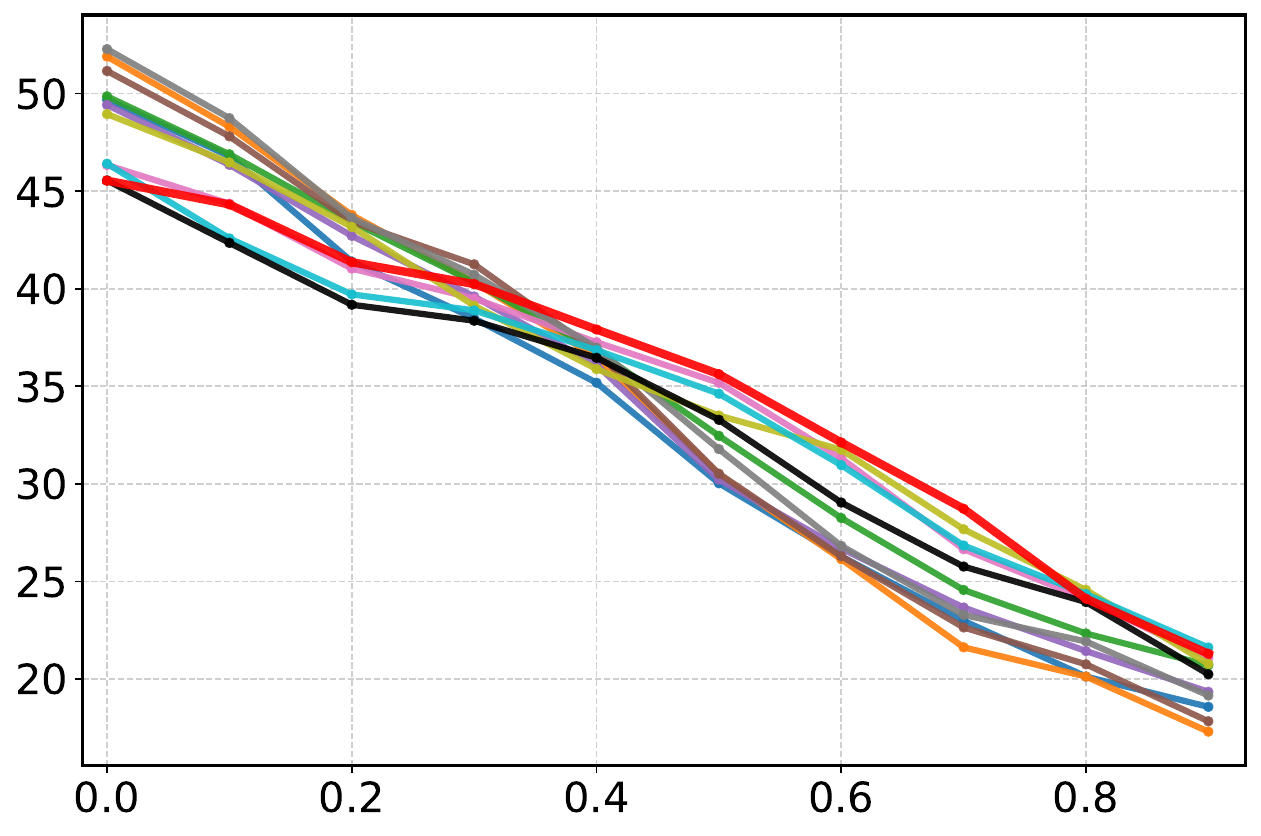} &
  \includegraphics[width=\imgwidth]{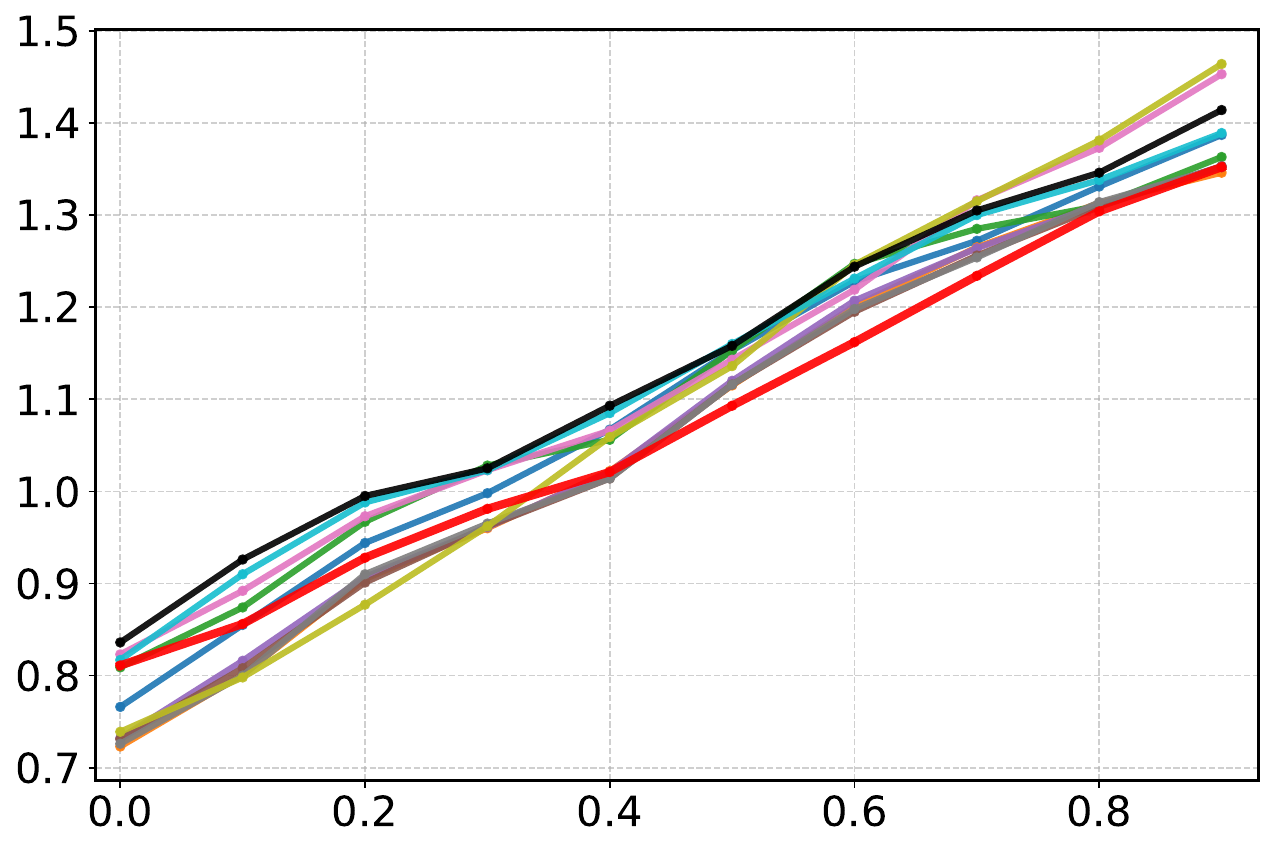} \\[-0.ex]

  \rowtitle{MOSEI} &
  \includegraphics[width=\imgwidth]{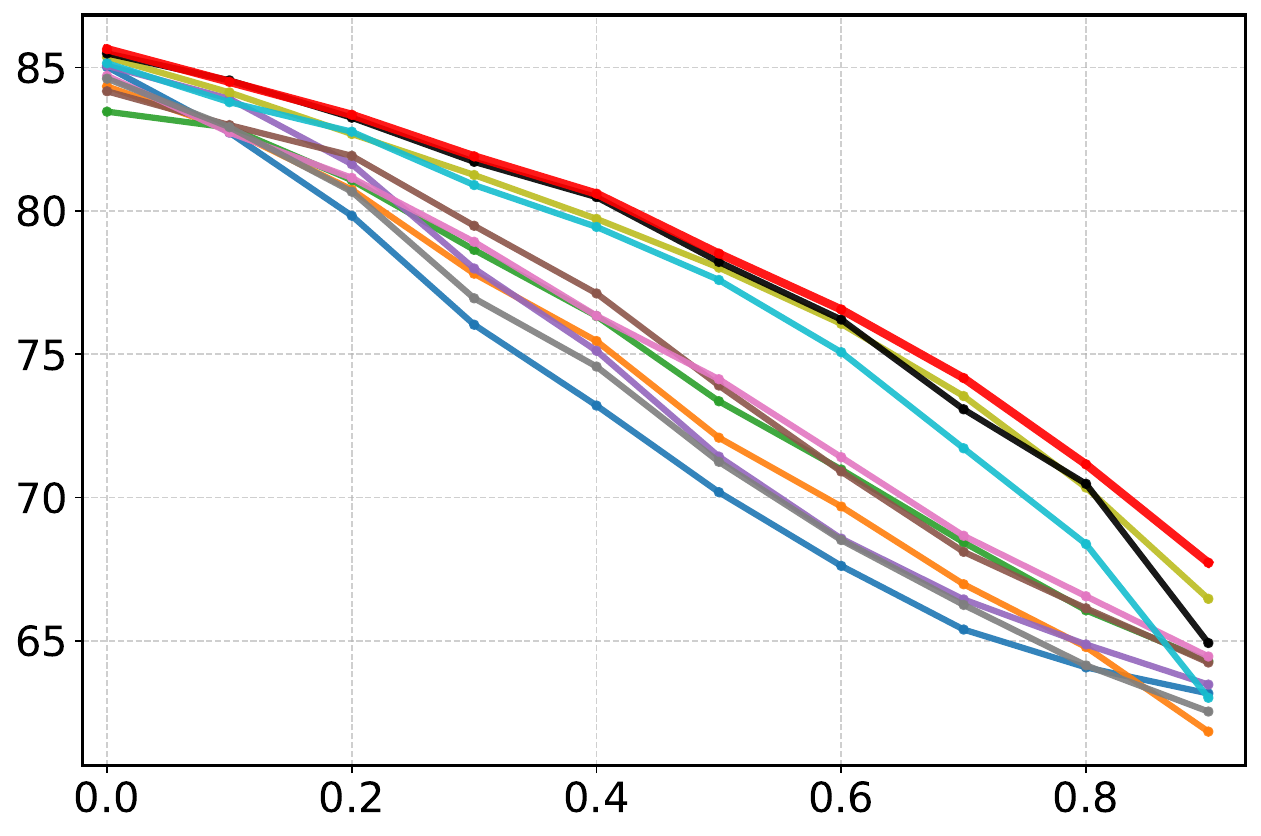} &
  \includegraphics[width=\imgwidth]{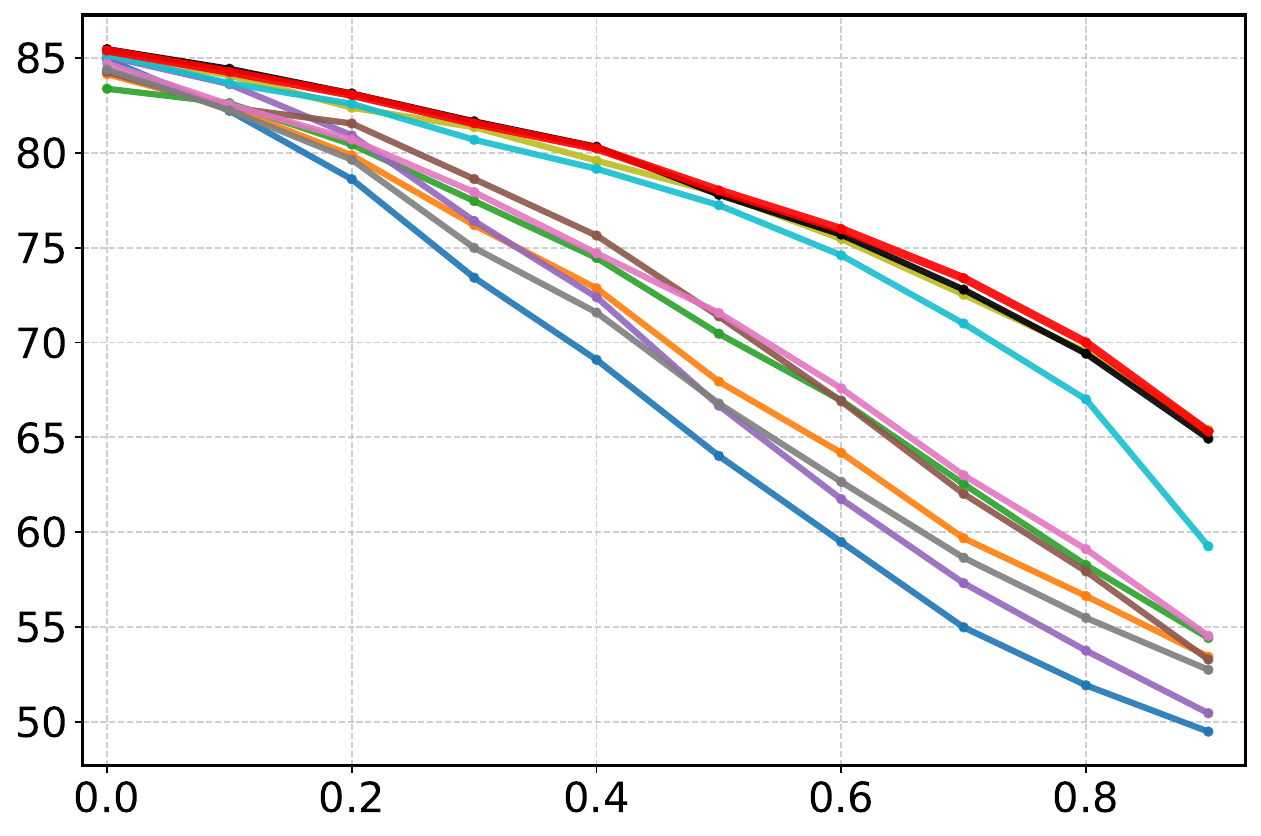} &
  \includegraphics[width=\imgwidth]{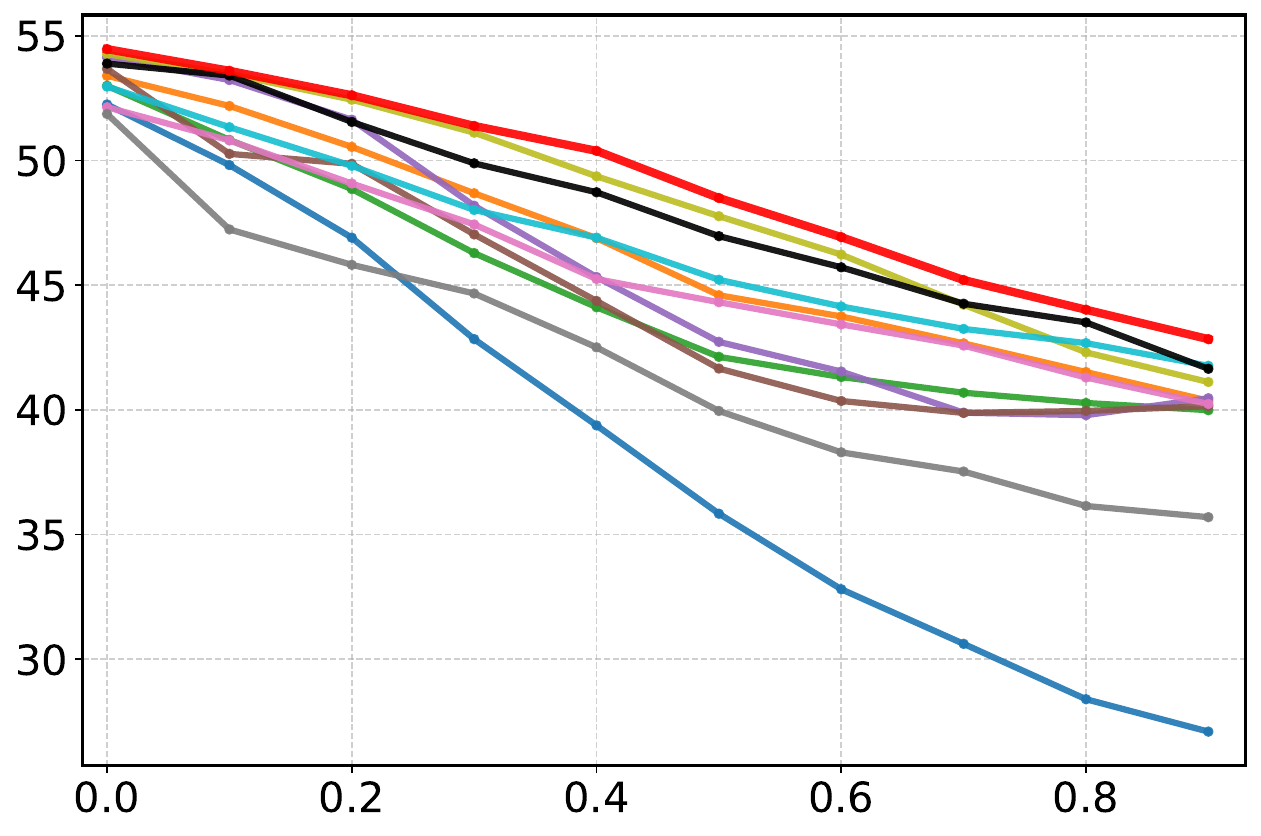} &
  \includegraphics[width=\imgwidth]{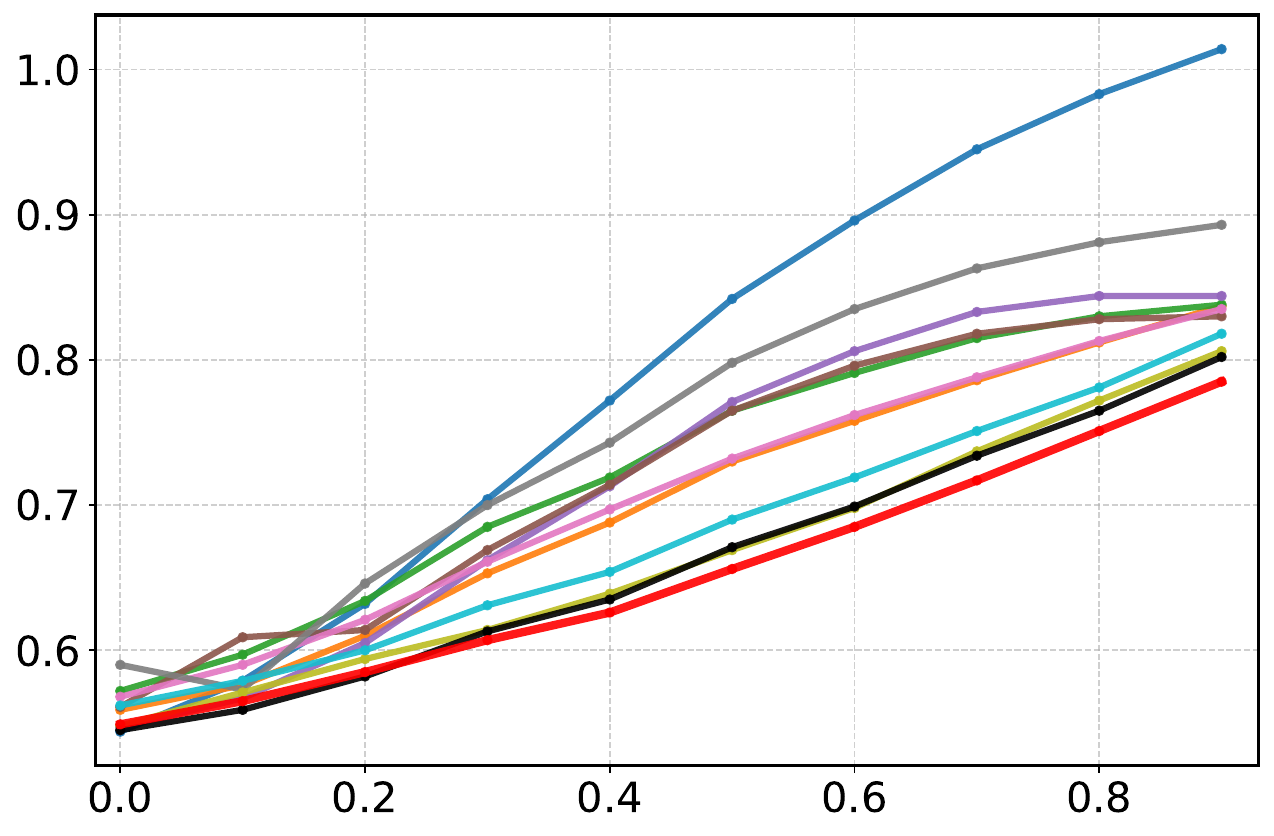} \\[-0.ex]

  \rowtitle{CH-SIMS} &
  \includegraphics[width=\imgwidth]{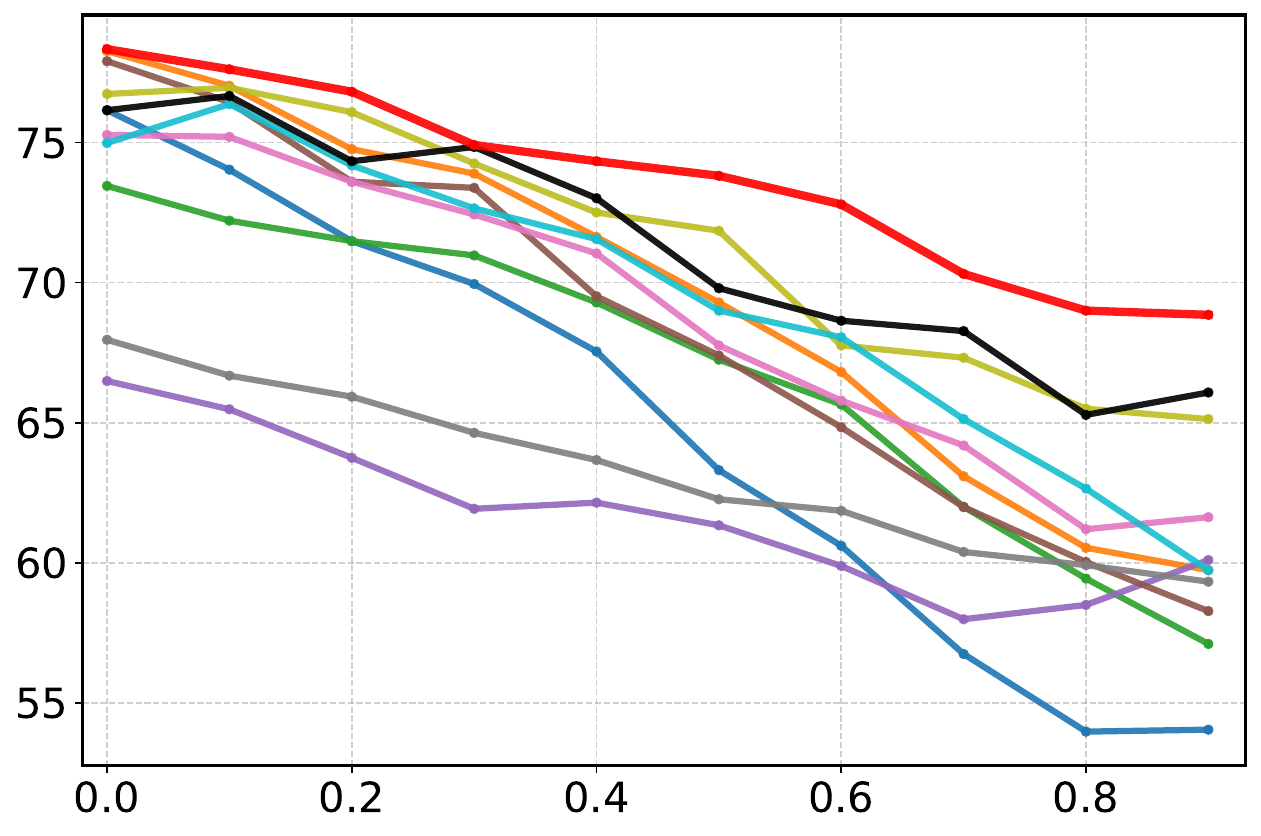} &
  \includegraphics[width=\imgwidth]{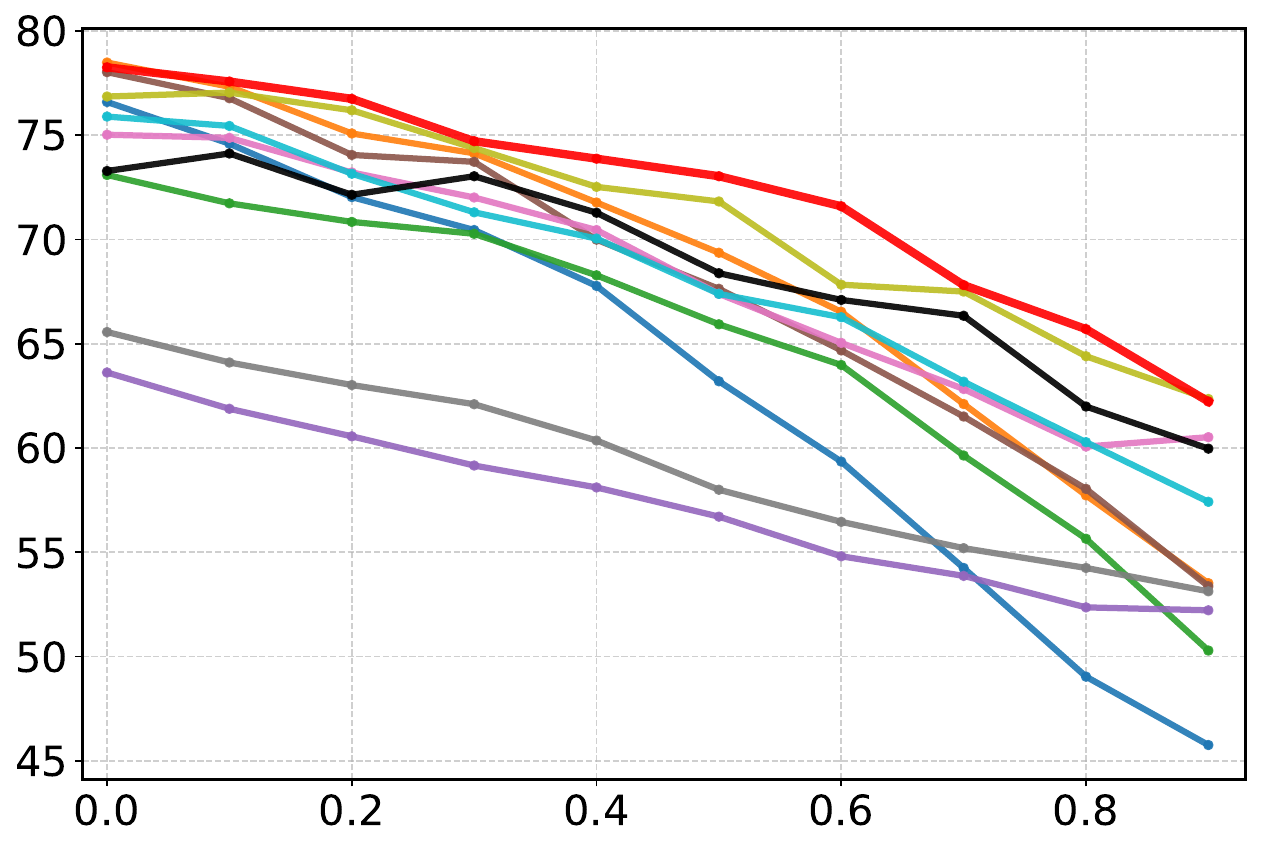} &
  \includegraphics[width=\imgwidth]{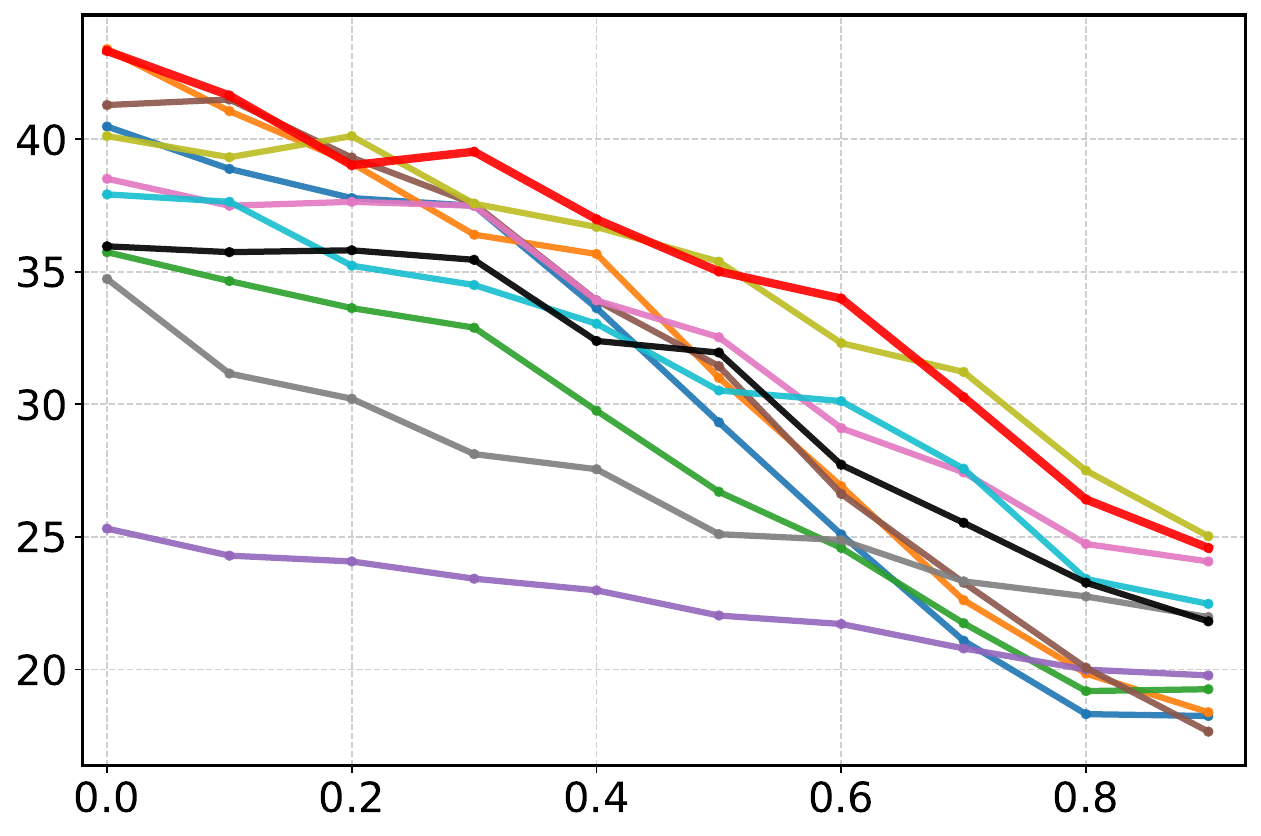} &
  \includegraphics[width=\imgwidth]{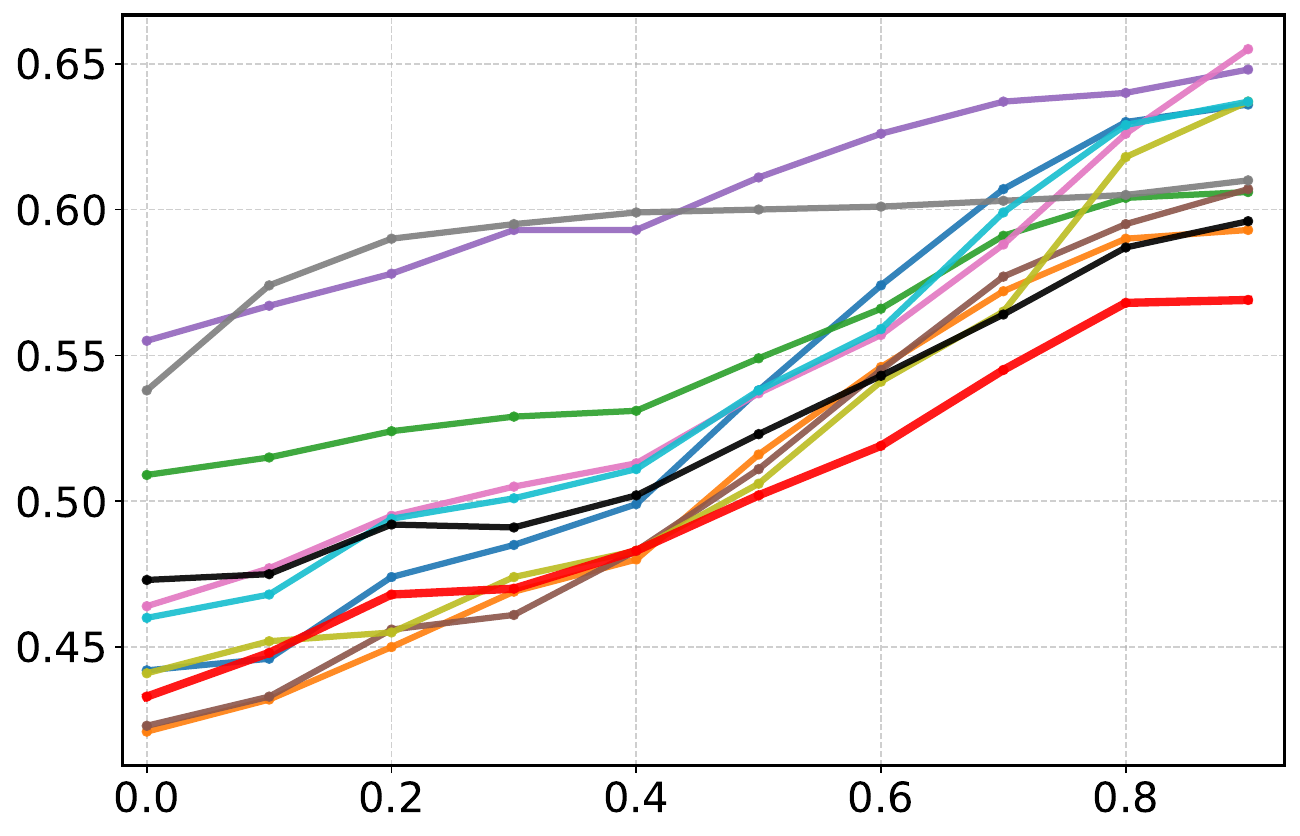} \\

  \\[-3.4ex]
  \multicolumn{5}{c}{%
    \includegraphics[width=0.94\textwidth]{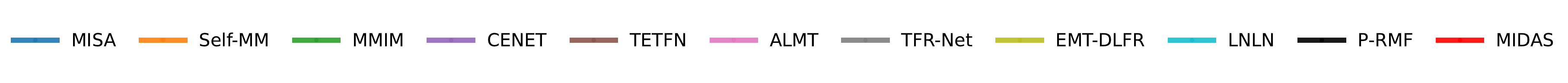}
  }
  
  \\[-1.8ex]
  \multicolumn{5}{c}{\scriptsize\textbf{Missing Rate $\boldsymbol{r}$}} \\

\end{tabular}%
}
\caption{Performance curves of varying missing rates on three datasets. Each row corresponds to a dataset (MOSI, MOSEI, CH-SIMS), and each column corresponds to a metric (Acc-2, F1, Acc-5, MAE). The X-axis represents the missing rate $r$, and the curves depict the performance of different methods. The legend at the bottom shows the color coding for the methods.}

\label{fig:6images}
\end{figure*}

To comprehensively evaluate robustness under real-world incomplete multimodal scenarios, we assess all models across a wide range of modality missing rates. For each evaluation metric, performance is averaged over all missing levels to provide an overall measure of robustness rather than sensitivity to a specific corruption ratio. The averaged results on MOSI and MOSEI are reported in Table~\ref{robust_comparison:mosi}, while the results on CH-SIMS are shown in Table~\ref{robust_comparison:sims}. In addition, performance trends across representative missing rates are illustrated in Figure~\ref{fig:6images}.

As shown in Table~\ref{robust_comparison:mosi} and Table~\ref{robust_comparison:sims}, MIDAS demonstrates strong and consistent improvements over existing approaches across all datasets, indicating its effectiveness in handling incomplete multimodal sentiment analysis. On the MOSI dataset, MIDAS achieves the best performance on most metrics, except for MAE. Specifically, it outperforms TFR-Net by 0.72\% in Acc-2 accuracy and improves the F1 score by 1.04\% over the strong EMT-DLFR baseline under the negative/positive setting. Under the negative/non-negative setting, it also achieves more than 1.0\% improvement over the best baseline. Although its MAE (1.074) is slightly higher than the lowest values achieved by Self-MM, TETFN and TFR-Net (1.065), MIDAS attains a substantially higher correlation score (0.534), suggesting more reliable sentiment intensity modeling.
On the MOSEI dataset, MIDAS consistently outperforms all competing methods across all evaluation metrics. Compared with the strong EMT-DLFR baseline, MIDAS achieves improvements of 0.77\% in Acc-5 and 0.52\% in Acc-7. Notably, MIDAS exhibits particularly strong performance in binary sentiment classification, with gains of 0.41\% in Acc-2 and 1.28\% in F1 score under the negative/non-negative setting. Furthermore, it achieves the lowest MAE (0.653) and the highest correlation score (0.607) among all compared methods, establishing a new state-of-the-art performance on MOSEI under incomplete conditions.

Performance gains are most pronounced on the CH-SIMS dataset, where MIDAS outperforms all baselines across every metric. Specifically, MIDAS achieves an Acc-2 of 73.68\% and an F1 score of 72.15\%, exceeding the strongest baseline EMT-DLFR by 2.27\% and 1.06\%, respectively. These results demonstrate the robustness of MIDAS in binary sentiment classification on more challenging multilingual and noisy data. In addition, MIDAS also achieves a 0.55\% gain in Acc-5 compared with EMT-DLFR, indicating more accurate modeling of fine-grained sentiment categories. It also attains the lowest MAE (0.501) and the highest correlation score (0.427), reflecting superior sentiment regression performance.
Figure~\ref{fig:6images} further illustrates performance trends across different missing rates. Compared with existing methods, MIDAS exhibits smoother and more stable performance degradation as modality missingness increases, highlighting its robustness and adaptability when a large portion of multimodal information is unavailable.

Several broader observations can be drawn from these results. First, robustness-oriented methods such as TFR-Net, EMT-DLFR, LNLN, P-RMF, and MIDAS consistently outperform traditional multimodal models under medium-to-high missing rates. While conventional methods may achieve competitive performance at low missing levels, their performance degrades rapidly as modality corruption intensifies, underscoring the importance of explicitly modeling robustness in practical applications. Nevertheless, some traditional models still produce competitive results on certain metrics, suggesting that components such as the Unimodal Label Generation Module (ULGM) in Self-MM and TETFN may serve as valuable building blocks for future robust MSA systems. Second, the advantage of MIDAS over existing incomplete MSA methods can be attributed to its explicit disentanglement of incomplete modality inputs into shared and exclusive latent representations, optimized through a mutual information minimax objective. This design enables MIDAS to better restructure corrupted inputs and maintain semantic alignment under severe degradation. Moreover, the uncertainty-aware fusion mechanism dynamically adjusts feature contributions, leading to more reliable multimodal integration.
Furthermore, MIDAS consistently outperforms MISA, a representative feature-decomposition approach, indicating that our principled MI-driven disentanglement is more effective than prior loss-based or constraint-based formulations under incomplete conditions. In particular, the joint supervision of prediction and reconstruction prevents the disentangled representations from collapsing into trivial commonalities or noise-dominated patterns, allowing the model to capture richer semantic information. Finally, we also observe that all methods perform better on the larger MOSEI dataset than on MOSI, suggesting that increased data scale and diversity can naturally enhance robustness in multimodal sentiment analysis.

\begin{table*}[t]
  \centering
  \renewcommand{\arraystretch}{}
  \caption{Effects of different components. Note: VM denotes Variational Modeling. MIM denotes MI Minimax. UAF denotes Uncertainty-aware fusion.}
  \label{ablation1}
  \begin{tabular}{lcccccccccccc}
    \toprule
    \multirow{2}{*}{Method}
      & \multicolumn{6}{c}{MOSI}
      & \multicolumn{6}{c}{CH-SIMS} \\
    \cmidrule(lr){2-7} \cmidrule(lr){8-13}
    & Acc-2 & F1 & Acc-5 & Acc-7 & MAE & Corr
    & Acc-2 & F1 & Acc-3 & Acc-5 & MAE & Corr \\
    \midrule
    \textbf{MIDAS} & \textbf{71.88}/\textbf{71.26} & \textbf{71.91}/\textbf{71.20} & 
                     35.12 & \textbf{32.06} &  \textbf{1.074} & \textbf{0.534}
                & \textbf{73.68} & \textbf{72.15} & 57.72 & \textbf{35.08} & 0.501 & \textbf{0.427} \\
    \midrule
    VM & 70.98/70.85 & 70.85/70.65 & 34.49 & 30.68 & 1.102 & 0.529
                & 72.45 & 71.43 & 57.43 & 33.23 & 0.505 & 0.409 \\
    VM+MIM & 71.50/71.06 & 71.59/70.96 & \textbf{35.32} & 31.81 & 1.078 & 0.533
                & 73.29 & 71.92 & 57.52 & 33.56 & \textbf{0.497} & \textbf{0.427} \\
    VM+UAF & 71.31/70.54 & 71.34/70.47 & 34.91 & 30.96 & 1.084 & 0.532
                & 72.72 & 71.51 & \textbf{57.78} & 33.11 & 0.513 & 0.409 \\
    \bottomrule
  \end{tabular} 
\end{table*}

\begin{table*}[t]
  \centering
  \renewcommand{\arraystretch}{1.}
  \caption{Effects of different regularizations. Note: $\mathcal{L}^{dis}$ represents disentanglement loss, $\mathcal{L}^{ali}$ represents alignment loss, $\mathcal{L}^{pred}$ represents prediction loss, and $\mathcal{L}^{rec}$ represents reconstruction loss.}
  
  \label{ablation2}
    \begin{tabular}{lcccccccccccc}
      \toprule
      \multirow{2}{*}{Method}
        & \multicolumn{6}{c}{MOSI}
        & \multicolumn{6}{c}{CH-SIMS} \\
      \cmidrule(lr){2-7} \cmidrule(lr){8-13}
      & Acc-2 & F1 & Acc-5 & Acc-7 & MAE & Corr
      & Acc-2 & F1 & Acc-3 & Acc-5 & MAE & Corr \\
      \midrule
      \textbf{MIDAS} & \textbf{71.88}/\textbf{71.26} & \textbf{71.91}/\textbf{71.20} & 
                     \textbf{35.12} & \textbf{32.06} &  \textbf{1.074} & \textbf{0.534}
                & \textbf{73.68} & \textbf{72.15} & 57.72 & \textbf{35.08} & \textbf{0.501} & \textbf{0.427} \\
      \midrule
      w/o \(\mathcal{L}^{dis}\)  & 70.59/70.26 & 70.37/70.02 & 34.23 & 30.56 & 1.110 & 0.524
               & 72.60 & 67.06 & 56.26 & 29.54 & 0.534 & 0.264 \\
      w/o \(\mathcal{L}^{ali}\)  & 70.84/70.56 & 70.58/70.29 & 34.19 & 30.55 & 1.095 & 0.532
               & 73.33 & 71.20 & \textbf{58.17} & 32.74 & 0.512 & 0.408 \\
      w/o \(\mathcal{L}^{pred}\) & 61.96/62.47 & 56.17/56.64 & 29.39 & 26.67 & 1.197 & 0.391
               & 72.41 & 66.09 & 57.05 & 29.39 & 0.533 & 0.317 \\
      w/o \(\mathcal{L}^{rec}\)     & 71.15/70.79 & 71.12/70.65 & 34.06 & 30.77 & 1.087 & 0.528
               & 72.94 & 71.93 & 57.24 & 33.32 & 0.508 & 0.419 \\
    \bottomrule
    \end{tabular}%
\end{table*}

\subsection{Effects of Different Components}

To evaluate the contribution of each component in our framework, we conduct comprehensive ablation studies on two representative datasets, MOSI and CH-SIMS, which cover different linguistic characteristics and cultural contexts. Specifically, we investigate the effects of three key components: variational modeling (VM), mutual information minimax optimization (MIM), and uncertainty-aware fusion (UAF).

We systematically remove or replace individual components while keeping the remaining architecture unchanged. As shown in Table~\ref{ablation1}, removing any component generally leads to performance degradation across most evaluation metrics on both datasets, although minor improvements on specific metrics can occasionally be observed. Such fluctuations are expected and can be attributed to two factors: (1) the inherent randomness introduced by incomplete multimodal inputs, and (2) the difficulty of tuning a single set of hyperparameters to simultaneously optimize all classification and regression metrics.

On the MOSI dataset, replacing the full MIDAS framework with VM alone leads to consistent drops in Acc-2, F1, and correlation, indicating that variational modeling by itself is insufficient to fully capture cross-modal dependencies under missing conditions. Incorporating MIM into VM (VM+MIM) noticeably improves Acc-2, F1, Acc-5, and Corr compared with VM, demonstrating that the mutual information minimax objective plays a crucial role in organizing incomplete multimodal information and strengthening semantic alignment. In contrast, combining VM with UAF (VM+UAF) yields more limited gains, suggesting that effective fusion alone cannot compensate for the lack of principled representation disentanglement.
Similar trends are observed on the CH-SIMS dataset. Compared to the baseline by VM, adding MIM consistently improves Acc-2, F1, MAE, and achieves the best correlation performance among the ablated variants. Although VM+UAF achieves a slightly higher Acc-3 score, it suffers from inferior overall classification and regression performance, highlighting the importance of disentanglement-driven representation learning over fusion-only enhancements. The full MIDAS model consistently achieves the best or near-best performance across all metrics, indicating that the three components are complementary rather than redundant.

Overall, these ablation results validate the effectiveness of each proposed component in our framework. In particular, the MIM module plays a central role in structuring incomplete multimodal inputs into meaningful shared and exclusive representations, while VM provides a stable probabilistic foundation and UAF enhances robustness during feature integration. Their combination enables MIDAS to learn robust and semantically aligned representations, resulting in strong and consistent performance under real-world incomplete multimodal scenarios.

\subsection{Effects of Different Regularizations}

To further examine the role of each regularization term in MIDAS, we conduct an additional set of ablation experiments by individually removing each regularization objective during training. The results on MOSI and CH-SIMS are reported in Table~\ref{ablation2}. Overall, removing any regularization component leads to noticeable performance degradation across most evaluation metrics, highlighting the importance of these objectives in stabilizing training and learning robust representations under incomplete multimodal settings.

Among all regularizations, the disentanglement loss $\mathcal{L}_{dis}$ and the alignment loss $\mathcal{L}_{ali}$ are particularly critical. Removing $\mathcal{L}_{dis}$ consistently degrades classification accuracy, MAE, and correlation on both datasets, indicating that explicit disentanglement is essential for separating shared and modality-exclusive information from corrupted inputs. Similarly, excluding $\mathcal{L}_{ali}$ leads to a substantial drop across almost all metrics, confirming its key role in enforcing semantic consistency among shared representations and preventing cross-modal misalignment.

We also observe a significant performance decline when removing the prediction loss $\mathcal{L}_{pred}$. This degradation is more pronounced than that caused by removing other regularization terms, since the model fails to converge under certain random seeds. We attribute this behavior to the fact that the optimization process becomes less constrained without direct predictive supervision. In the absence of $\mathcal{L}_{pred}$, the latent representations are guided only by auxiliary objectives, which may result in unstable optimization dynamics and suboptimal local minima. This phenomenon highlights the importance of prediction-level supervision in anchoring representation learning and ensuring stable convergence.

In contrast, removing the reconstruction loss $\mathcal{L}_{rec}$ results in relatively moderate performance drops on most metrics. However, it still causes noticeable degradation in fine-grained classification performance, such as Acc-7 on MOSI and Acc-5 on CH-SIMS. This suggests that reconstruction-based regularization helps recover missing or corrupted information and provides additional structural constraints that are particularly beneficial for modeling subtle sentiment intensity variations.

Overall, these results demonstrate that the proposed regularization terms are complementary rather than redundant. While $\mathcal{L}_{dis}$ and $\mathcal{L}_{ali}$ are primarily responsible for learning structured and semantically aligned representations, $\mathcal{L}_{pred}$ provides a semantic anchor that stabilizes optimization, and $\mathcal{L}_{rec}$ further improves robustness by encouraging information recovery. Their joint optimization is therefore essential for achieving strong and consistent performance in incomplete MSA.

\textbf{Training Stability and Co-optimization}. Optimizing the multi-term objective raises natural questions regarding potential gradient conflicts and convergence stability. Empirically, we observe that MIDAS converges smoothly without severe instability. This is because the five distinct objectives act in a complementary rather than adversarial manner. The auxiliary constraints $\mathcal{L}^{dis}$, $\mathcal{L}^{ali}$, and $\mathcal{L}^{rec}$ effectively shape and regularize the latent space, while the auxiliary prediction loss $\mathcal{L}^{pred}$ acts as a semantic anchor. This anchoring prevents the gradients of the information-theoretic objectives from driving the representations into trivial or unstable regions. Furthermore, the explicit use of a learning rate warm-up strategy combined with a cosine annealing scheduler mitigates early-stage gradient variance, ensuring stable co-optimization and consistent convergence of the primary task loss $\mathcal{L}^{task}$.

\begin{figure*}[ht]
  \centering
  \includegraphics[width=0.96\textwidth]{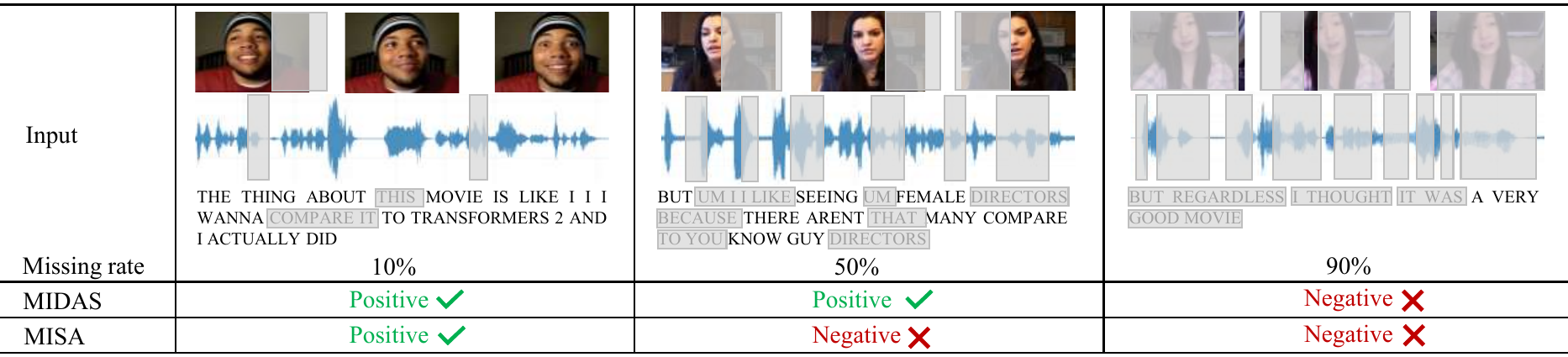}
  \caption{Visualization of example predictions by MIDAS and MISA under varying missing rates (10\%, 50\%, and 90\%) from the MOSI dataset. Note: The input is visualized to facilitate readers’ understanding. In practice, random masking is applied to the original input sequence.}
  \label{fig:case}
\end{figure*}

\begin{table}[t]
\caption{Analysis on model complexity. Note: the parameter of other methods was calculated from open source code with default hyper-parameters on MOSI without pre-trained BERT. M and G denote million and billion, respectively.}
  \centering
  \renewcommand{\arraystretch}{1}  
  \begin{tabular}{lccc}
    \toprule
    Method & Params(M) & FLOPs(G) & F1 \\
    \midrule
    \textbf{MIDAS} & \textbf{1.23} & \textbf{17.53} & \textbf{71.91/71.20}  \\
    \midrule
    TFR-Net & 7.12 & 19.74 & 69.39/67.58  \\
    EMT-DLFR & 1.81 & 18.56 & 69.39/67.58  \\
    LNLN & 5.96 & 19.47 & 70.01/69.49  \\
    P-RMF & 7.31 & 19.51 & 69.82/68.53 \\
    \bottomrule
  \end{tabular}
  \label{ablation3}
\end{table}

\subsection{Efficiency Analysis}

As shown in Table~\ref{ablation3}, we compare MIDAS with representative incomplete MSA methods in terms of model size, computational complexity, and F1 performance on the MOSI dataset. For a clear comparison of model design, the reported parameter counts exclude the pre-trained BERT encoder (110M).
MIDAS achieves the best F1 scores (71.91/71.20) while using only 1.23M parameters, which is substantially fewer than those of TFR-Net (7.12M), EMT-DLFR (1.81M), LNLN (5.96M), and P-RMF (7.31M). In addition to its compact parameterization, MIDAS also exhibits the lowest computational cost, requiring only 17.53G FLOPs per forward pass, compared with approximately 18.5-19.5G FLOPs for the competing methods.

Despite its lightweight design, MIDAS consistently outperforms larger models in terms of classification performance, demonstrating a favorable trade-off between model capacity, computational efficiency, and effectiveness. These results indicate that the proposed MI-driven representation learning framework enables strong sentiment modeling capability without relying on large parameter budgets or heavy computational overhead, while further promoting robust generalization across diverse incomplete multimodal scenarios.

\subsection{Case Study}

As illustrated in Figure~\ref{fig:case}, we visualize several predictions made by MIDAS and MISA under different missing rates (10\%, 50\%, and 90\%) from the MOSI dataset. At a low missing rate of 10\%, the multimodal input retains sufficient semantic information, allowing both MIDAS and MISA to successfully output accurate positive predictions. However, as the missing rate increases to 50\% and significant portions of the text, audio, and visual cues are corrupted, MISA fails and incorrectly predicts a negative sentiment. In contrast, MIDAS maintains its robustness and correctly identifies the positive sentiment. This stark contrast highlights MIDAS’s superior ability to capture and leverage latent task-relevant semantics in highly incomplete scenarios, which is facilitated by its explicit disentanglement and uncertainty-aware fusion mechanisms. Finally, under an extreme missing rate of 90\%, both models fail to output correct predictions. We attribute this to the catastrophic degradation of valid information across all multimodal inputs, which effectively erases the core semantic content and renders accurate sentiment analysis nearly impossible.

\section{Conclusion and Future Work}

In this article, we present a unified framework, termed Mutual Information Disentanglement with uncertainty-Aware fuSion (MIDAS), which aims to effectively extract and leverage task-relevant semantics for robust multimodal sentiment analysis under incomplete data conditions. MIDAS integrates variational latent modeling with an information-theoretic disentanglement strategy, restructuring corrupted multimodal inputs into shared and exclusive latent spaces that explicitly decouple modality-invariant semantics from modality-exclusive factors. By minimizing intra-modal mutual information, the framework promotes clean and noise-resistant disentanglement; by maximizing cross modal mutual information, it strengthens semantic alignment across modalities. Moreover, the proposed uncertainty-aware fusion mechanism leverages variational uncertainty estimates to adaptively weight and integrate modality contributions, enabling reliable predictions even when certain modalities are incomplete. Extensive experiments on benchmark datasets demonstrate that MIDAS consistently outperforms state-of-the-art baselines and exhibits strong resilience under diverse incomplete-data scenarios.

Despite these promising results, real-world multimodal data exhibit more complex and dynamic missingness patterns than those captured in controlled benchmarks. In practice, different modalities often suffer from asymmetric or uneven missing proportions. Future research will explore strategies for modeling realistic, temporally evolving missingness behaviors and developing adaptive fusion mechanisms capable of generalizing across these unpredictable conditions. In parallel, establishing standardized evaluation protocols for heterogeneous missing scenarios remains both challenging and essential. Furthermore, extending MIDAS to large-scale pretraining regimes and exploring its applicability to broader multimodal tasks, such as emotion recognition and human-computer interaction, constitutes promising directions for future research.

\bibliographystyle{IEEEtran}
\bibliography{main}

\vfill

\end{document}